\documentclass{article}

\usepackage[final]{corl_2019} % Uncomment for the camera-ready ``final'' version

\newcommand{\figref}[1]{Fig.\ref{#1}}

\usepackage{amsmath}
\usepackage{amssymb}
\newcommand{\sS}{\mathbf{s}}
\newcommand{\bS}{\mathbf{h}}
\newcommand{\bSp}{\mathbf{h}^\prime}
\newcommand{\bO}{\mathbf{o}}
\newcommand{\bA}{\mathbf{a}}
\newcommand{\VE}{\hat{V}}
\newcommand{\VM}{\tilde{V}}
\newcommand{\rE}{\hat{r}}

\usepackage{multicol}
\usepackage{graphicx}
\usepackage{subcaption}
\usepackage{algorithm}
\usepackage{algorithmic}

\newlength\myindent
\newcommand\bindent{%
  \begingroup
  \setlength{\itemindent}{\myindent}
  \addtolength{\algorithmicindent}{\myindent}
}
\newcommand\eindent{\endgroup}

\usepackage[normalem]{ulem}

\title{Imagined Value Gradients: Model-Based Policy Optimization with Transferable Latent Dynamics Models}
\author{
  Arunkumar Byravan$^2$\thanks{Work done during an internship at DeepMind.}
  \ \ \ \ \textbf{Jost Tobias Springenberg}$^1$
  \ \ \ \textbf{Abbas Abdolmaleki}$^1$
  \ \ \ \textbf{Roland Hafner}$^1$
  \And Michael Neunert$^1$
  \And Thomas Lampe$^1$
  \And Noah Siegel$^1$
  \And Nicolas Heess$^1$
  \And Martin Riedmiller$^1$
  \and $^1$ DeepMind, London
  \and $^2$ University of Washington
}
\date{May 2019}

\begin{document}
\maketitle
\vspace{-0.5cm}
%===============================================================================
\begin{abstract}
Humans are masters at quickly learning many complex tasks, relying on an approximate understanding of the dynamics of their environments. In much the same way, we would like our learning agents to quickly adapt to new tasks. In this paper, we explore how  model-based Reinforcement Learning (RL) can facilitate transfer to new tasks. We develop an algorithm that learns an action-conditional, predictive model of expected future observations, rewards and values from which a policy can be derived by following the gradient of the estimated value along imagined trajectories. We show how robust policy optimization can be achieved in robot manipulation tasks even with approximate models that are learned directly from vision and proprioception. We evaluate the efficacy of our approach in a transfer learning scenario, re-using previously learned models on tasks with different reward structures and visual distractors, and show a significant improvement in learning speed compared to strong off-policy baselines. Videos with results can be found at \url{https://sites.google.com/view/ivg-corl19}

%Finally, we show transfer results on finetuning previously learned models on tasks with different reward structures and visual distractors; we also discuss performance improvements when a few demonstrations from the target task are available.

%We’d like to see how models help model-free algorithms learn faster. Mainly, we focus on using representations learned from a model as the state representation for model-free algorithms and how this helps speed up learning, allows for faster transfer across tasks and improves robustness of the learned policies.
\end{abstract}


%%%%%%%% Split into files
%===============================================================================
\section{Introduction}
\vspace{-0.2cm}
The last decade has seen significant progress in reinforcement learning. The field has matured to a state where RL can solve challenging simulated motor control problems \cite{heess2017emergence, openai2018} or games  \cite{silver2018general,alphastarblog} even from high-dimensional observations such as images \cite{mnih2015human, mbrl-atari,espeholt2018impala}. 
Off-policy model-free algorithms have become workable for high-dimensional continuous action spaces %and motor control tasks
\cite{lillicrap2015continuous,heess2015learning,abdolmaleki2018maximum}, and have improved in robustness and data-efficiency allowing experiments directly on robotics hardware
\cite{abdolmaleki2018relative,haarnoja2018soft,zhu2018, kalashnikov2018qt}.

Model-free techniques directly learn a policy (and value function) from environment interactions. Their simplicity, generality and versatility has been a major factor behind their recent successes. 
%\akb{Does not flow well (in the rest of the para) from here on}
Yet, these techniques do not entirely satisfy the intuition that a learning agent should be able to acquire approximate knowledge of the dynamics of its environment in a manner that is independent of any particular task and such that it is easily applicable to new tasks, more easily than task-specific objects such as policies. It is this intuition and as well as the desire to further improve the sample-efficiency and transferability of solutions that has driven much of the recent research in model-based RL.
% %, and be able to leverage this knowledge to improve its behavior. 
% Such knowledge would then be transferable to new tasks in the same domain more effectively than task-specific objects such as policies. It is this intuition and, more practically speaking, the desire to further improve on the sample-efficiency and transferability of model-free solutions that has driven much of the recent research in model-based RL.

A growing body of literature is concerned with learning dynamics models for physical control problems \cite{deisenroth2011pilco,visual-foresight}, including approaches that learn latent models from images \cite{watter2015embed,planet-dm,solar-icml19}. Although some approaches excel in data efficiency  \cite{deisenroth2011pilco,levine2016end}, in general, model-based methods have not yet achieved the robustness of model-free techniques; and they still struggle with model inaccuracies and long planning horizons. When learned dynamic models are combined with policy learning \cite{sutton1990integrated,gu-modelbasedacc,imagination-dm,mve-feinberg,wayne2018unsupervised} the advantages over purely model-free techniques can be less clear.

%In some cases it has been possible to achieve very data efficient control with such approaches \cite{XYZ,ABCTODO}. 
%In general, such model-based approaches have, however, not yet been able to achieve the same robustness and versatility as model-free techniques due to problems in the presence of model inaccuracies. 

%\tobi{removed the paragraph on lots of related work here as I think we have this down in the related work already it is only commented out though}
%There has been a number of works exploring learned dynamics models for physical control problems \cite{XXX} including approaches that employ latent dynamic models and are hence capable to operate in domains when the system is observed e.g. through images \cite{XYZ}. Although in some cases it has been possible to achieve very data efficient control \cite{XYZ}, these approaches often dispense of a policy and rely on model predictive control (i.e.~rely on planning through the model) at test time. This can, however, lead to problems in the presence of model inaccuracies, or when a task is defined by sparse rewards or requires a long planning horizon. More generally, such model-based approaches have not yet been able to achieve the same robustness and versatility as model-free techniques. 
% planning horizon, model quality
%\akb{This is isolated, maybe start with saying there has been work in the intersection but advantages are not clear?}
%There has also been work combining learned dynamic models with policy learning \cite{xxx}, however, in such cases the advantages over purely model-free techniques are often less clear. 

In this work we develop an approach that uses model-based RL to learn a stochastic parametric policy in domains with high-dimensional observation spaces such as images. We build on Stochastic Value Gradients (SVG) \cite{heess2015learning} which allow us to compute low-variance policy gradients by directly differentiating a model-based value function.
We extend this work in two ways: Firstly, we develop a latent state space model that allows us to predict expected future reward and value as a function of the current policy even when the true low-dimensional system state is not directly observed. Secondly, rather than using the model only for credit assignment along observed trajectories we use it directly to produce ``imagined'' rollouts and show that stable policy learning can still be achieved.

We apply our algorithm to several challenging long-horizon vision-based manipulation tasks (e.g.~lifting and stacking) in simulation and demonstrate the following:
Our model-based approach (a) is as robust and achieves the same asymptotic performance as competitive model-free baselines, and (b) in several cases it significantly improves data-efficiency. (c) It can effectively transfer the learned model to novel tasks with different reward distributions or visual distractors, leading to a dramatic gain in data-efficiency in such settings. (d) It is particularly effective in a multi-task setup where the models learned on multiple tasks learn faster and generalize better to new tasks, successfully solving problems that cannot be learned in isolation.

\vspace{-0.2cm}
\section{Related Work}
\vspace{-0.1cm}
%\akb{Do we need this section? A lot of this is discussed above in the intro.}\\
%\akb{We can add a section on latent space models here if we decide to focus more on models + transfer. Might make more sense.}
%%%%%%%%%%%%%%%%%%%%%%%%
\textbf{Model-free RL:} Model-free RL has recently been applied to many challenging problems including robotics \cite{deisenroth2013survey}. These successes were
helped by advances in algorithms suitable for the use of powerful function approximators \cite{mnih2015human, lillicrap2015continuous, schulman2017proximal, abdolmaleki2018maximum, silver2018general, riedmiller2018learning}, also enabling the use of RL in domains with high-dimensional observation spaces \cite{mnih2015human, alphastarblog, riedmiller2018learning}.
However, model-free methods can still struggle in the low data regime and their primary objects of interest, policies and value functions are task specific and can thus be difficult to transfer.

%%%%%%%%%%%%%%%%%%%%%%%%
\textbf{Model-based RL:} On the other end of the spectrum, model-based methods learn a model of the environment and use a planner to generate actions. One of the successes in this area is work by Deisenroth et al. \cite{deisenroth2011pilco} who learn uncertainty aware policies using Gaussian-Process models and showed impressive data efficiency on multiple low-dimensional tasks. Using deep neural networks, model based RL has been extended to vision-based problems such as Atari \cite{mbrl-atari} and complex manipulation tasks \cite{levine2016end, visual-foresight}. Other work has studied learned vision-based latent representations for RL \cite{watter2015embed, wahlstrom2015pixels, solar-icml19, cobra-dm, planet-dm}: Such low-dimensional representations 
%can provide good, task-specific, 
can provide feature spaces in which learning and planning can proceed significantly faster compared to learning directly from images 
% However, problems
% %While these approaches generate impressive results, they often struggle 
% with model inaccuracies and on tasks with sparse rewards and long-horizons remain.
although model inaccuracies, sparse rewards, and long-horizons remain challenging.

%\cite{visual-foresight} learns a video prediction model based on self-supervised training data, primarily generated through random exploration followed by CEM style planning with different reward function specifications (pixel error, goal images and goal classifiers).

%\citep{planet-dm} is another model-based RL method that learns a latent state representation from pixels and uses the CEM for planning in the learned latent space. The learned model has both stochastic and deterministic components and is trained to minimize long-term prediction error. The method is tested on visual control tasks in mujoco.

%\citep{solar-icml19} is a purely model-based RL method that learns a state representation from images that can be used in conjunction with iLQR style controllers (similar to E2C). The approach is tested on real world stacking tasks where the block is already grasped \& the target position is fixed. It is also trained on a real-world pushing task.

%\citep{ke2018modeling} is also a model-based RL approach that learns a predictive model with an auxiliary loss for consistent long-term future predictions. This mode is then used in conjunction with an MPC style planner and a PPO style optimization algorithm for learning exploration policies. It is tested on mujoco tasks with visual observations.

%\citep{cobra-dm} from DeepMind learns a representation along with a 1-step transition model, exploration policy and reward predictor for simple 2D object movement tasks. Uses the MONet architecture for representation learning. \\

%%%%%%%%%%%%%%%%%%%%%%%%
\textbf{Model-free + Model-based:} Several papers have focused on combining the strengths of model-free and model-based RL. The work on Dyna \cite{sutton1990integrated} was among the first. It integrates an action model and model-based imagined rollouts with policy learning, interleaving planning, learning and execution in a tight loop. Recent work has further explored the use of imagined rollouts generated with learned models for accelerating learning of model-free policies \cite{gu-modelbasedacc, imagination-dm} -- or indirectly speeding up learning via model-based value estimates \cite{mve-feinberg, steve-nips19}. These approaches propose different means to handle model approximation errors, such as using short rollouts to avoid cascading model errors \cite{mve-feinberg}, uncertainty estimation through model ensembles \cite{steve-nips19} and using the rollouts as policy conditioning variables only \cite{imagination-dm}. Contrasting these methods, Stochastic Value Gradients (SVG) \cite{heess2015learning} re-evaluates rollouts with a learned model from off-policy data, accelerating learning through value gradients back-propagated through time via a learned model. In this work, we extend SVG to use imagined rollouts in latent spaces, accounting for model approximation errors via gradient averaging.

%\textbf{Models and Latent Representations:} With the advent of convolutional neural networks, there has been significant progress on learning image-based models and representations for planning and reinforcement learning \cite{ebert2018visual, cobra-dm, byravan2017se3}. 

There has also been work on combining latent space models and model-free RL. DeepMDP \cite{gelada2019deepmdp}, MERLIN \cite{wayne2018unsupervised}, CRAR \cite{franccois2018combined} and VPN \cite{oh2017value} combine representation learning with policy optimization via auxiliary losses such as observation reconstruction \cite{wayne2018unsupervised}, predicting the next latent state \cite{gelada2019deepmdp} or future values \cite{oh2017value}. Along these lines, we learn a latent representation that can be used for predicting (expected) future latent states, observations, rewards and values conditioned on the observed actions.

\textbf{Transfer:}
One argument for model-based RL is the potential of transfer to tasks in the same environment. Sutton et al. \cite{sutton1990integrated} discuss early examples of this type of transfer on simple problems. Recently, Francois et al. \cite{franccois2018combined} showed encouraging results with a vision-based RL agent on 2D labyrinth tasks. Nagabandi et al. \cite{nagabandi2018learning, nagabandi2018deep} meta-learn predictive dynamics models to enable a model-based agent to rapidly adapt to changes in its environment. Other work on transfer in RL has focused on learning reusable skills e.g.\ in the form of embedding spaces \citep{hausman2018learning,florensa2017,dhruva2019}, successor representations \citep{barreto2017}, transferable priors \citep{galashov2018} or meta-policies \citep{finn2017,clavera2018}. As they learn ``behaviors'' rather than dynamics models, they are in a sense orthogonal and complementary to the ideas presented in this work.

%===============================================================================
\vspace{-0.2cm}
\section{Background}
\vspace{-0.2cm}
%\akb{I have structured this similar to the VPN paper and a lot of the math/MDP background stuff is borrowed from the SVG \& VPN papers. Change it if it is too similar to them.}\\

We are interested in solving motor control problems such as robotic manipulation tasks from vision. 
This setup can be formalized as a partially observed Markov decision process (a POMDP) with observation $\bO^t \in R^{N_O}$, states $\sS^t \in R^{N_S}$, actions $\bA^t \in R^{N_A}$ 
%TOBICOPY The latent state space is constructed via a deterministic mapping from the observations: $\bS^t = f_{\text{enc}}(\bO^{1:t})$ To obtain low-variance estimates in the policy optimization, we will consider systems that can be described with deterministic latent state spaces (although the observations could, in principle, be stochastic realizations). 
transition probabilities $p(\sS^{t+1} | \sS^{t}, \bA^t)$,
%= p(\sS^{t+1} | \bO^{1:t}, \bA^t)$. 
and reward function $r^t = r(\sS^t, \bA^t)$. %This MDP mirrors the ``true environment'' MDP in the observation space $O$: $\bar{\mathcal{M}} = (O, A, \bar{p}, \bar{r})$, sharing the same action space $A$ with the transition probability distribution $\bar{p}(o^{t+1} | o^{1:t}, a^t)$ and reward $\bar{r}$ functions operating on observations $\bO^t$. 
Let $\tau_{\leq t} = (\bO^{1:t}, \bA^{1:t-1})$ be the sequence of observations and actions in a trajectory up to decision point $t$.
The optimal policy and the value function at time step $t$ in this setting are a function of the posterior of the system state $p(\sS_t | \tau_{\leq t})$ given the interaction history. However, this posterior is usually intractable and many reinforcement learning approaches resort, instead, to directly optimizing a parametric policy that is a function of the history $\tau_{\leq t}$, i.e.\ they consider policies of the form $\pi(\bA_t | \tau_{\leq t})$.
Thus, they aim to maximize the sum of discounted rewards: $J(\theta) = \mathbb{E}_{p_\pi(\sS^{t:T}, \bA^{t:T}, \bO^{t+1:T} | \tau_{\leq t})}\left[ \sum_{t=0}^T \gamma^t r^t\right]$, with discount factor $\gamma \in [0, 1)$, where the trajectory distribution is assumed to decompose into transition and action probabilities as 
$
p_\pi(\sS^{t:T}, \bA^{t:T}, \bO^{t+1:T} | \tau_{\leq t}) = p(\sS^t | \tau_{\leq t}) \prod_{t' \geq t}^{T-1} \pi(\bA^{t'} | \tau_{\leq t'}) p(\sS^{t'+1} | \sS^{t'}, \bA^{t'}) p(\bO^{t'+1} | \sS^{t'+1}).
$
This allows us to define the value of an observed trajectory prefix as 
\begin{equation}
\textstyle
%V^\pi(\sS^t) =
V^\pi(\tau_{\leq t}) = \mathbb{E}_{p_\pi(\sS^{t:T}, \bA^{t:T}, \bO^{t+1:T} | \tau_{\leq t})} \left[ \sum_{k=t}^T \gamma^{k-t} r^t 
%| \sS^t = \sS 
\right],
\label{eqn:true_v}
\end{equation}
where, below, we consider the  infinite horizon case" $\lim_{T \to \infty}$.

\vspace{-0.2cm}
\section{An Action-Conditional Expectation Model of Observations and Rewards}
\vspace{-0.2cm}

Naively, a model-based evaluation of \eqref{eqn:true_v} would require an accurate action-conditional model of future observations and rewards given a a partial trajectory $\tau_{\leq t}$. Especially in high-dimensional observation spaces such a model can be difficult to learn.
%
%
% We are interested in a model based RL solution for finding $\pi$ in the scenario described above. Unfortunately, modeling the distributions comprising %$p_\pi(\sS^{t:T}, \bA^{t:T}, \bO^{t+1:T} | \tau_{\leq t})$
% $p_\pi(r^{t:T}, \bA^{t:T}, \bO^{t+1:T} | \tau_{\leq t})$
% -- to estimate Equation \eqref{eqn:true_v} for model based RL -- can be difficult  \citep{TODONicolasadd} that as it involves fitting a probabilistic models of all potential future observations, and rewards, given a partial trajectory. 
%
Instead we consider an approximate model suitable for weakly partially observed domains with limited stochasticity. 
%(We discuss model limitations in Sec.~\ref{sec:model:limitations}.) 
We develop a latent state-space model whose latent state at time step $t$ is optimized to represent a sufficient statistic of the interaction history $\tau_{\leq t}$. The model is trained to predict expectations of future observations and reward by approximately modeling the evolution of the summary statistic via a deterministic transition model.
%As a simplification we instead fit a latent state-space model whose latent state at time step $t$ is optimized to represent a sufficient statistic of the interaction history $\tau_{\leq t}$. The model is trained to predict expectations of future observations and reward by approximately modeling the evolution of the summary statistic via a deterministic transition model.
%
%As a simplification we instead fit a latent state-space model whose latent state at time step $t$ is optimized to represent a sufficient statistic of the interaction history $\tau_{\leq t}$. The model is trained to predict expectations of future observations and reward by approximately modeling the evolution of the summary statistic via a deterministic transition model.
%
We express the policy as a function of the latent state and the model allows us to construct a surrogate model $\VM$ that expresses the value $V^\pi(\tau_{\leq t})$ as a recursive function of the policy, the deterministic transition function, and a learned approximation to the reward. This surrogate model can be used to compute approximate policy gradients.

More specifically, we let $\bS^t = f_\text{enc}(\bO^{1:t}; \phi)$ be a determistic mapping (with parameters $\phi$) extracting a summary statistic from 
%latent state from 
a history of observations\footnote{We drop the dependency on actions here, assuming $\sS$ is retrievable from a history of observations only.}; and we let $\bS^{t+1} \triangleq f_{\text{trans}}(\bS^t, \bA^t; \phi)$ denote a latent transition function. %As noted above and discussed in more detail below (\ref{XXX}) we make the simplifying assumption that  any expectation of future observations or rewards can be represented as a deterministic function of $\bS^t$ conditioned on future actions. 
The assumption that the latent dynamics are well described by a deterministic transition function is our primary simplification.
We further define the approximate reward function based on latent states as $\rE(\bS^t, \bA^t; \phi) = \rE( f_\text{trans}(\bS^t, \bA^t); \phi)$ which we calculate by chaining the transition model with a reward predictor. Finally, we denote with $\pi_\theta(\bA | \bS)$ a 
stochastic policy, parameterized by $\theta$. Using these definitions we construct an approximation of the expectation in Eq. \eqref{eqn:true_v} as $V^\pi(\tau_{\leq t}) \approx \VM^\pi(\bS^t) = \mathbb{E}_{\pi_\theta} \left[ \sum_{k=t}^\infty \gamma^{k-t} \rE^k | \bS^t = \bS \right]$ or, written recursively:
\begin{equation}
    V^\pi(\tau_{\leq t}) \approx \VM^\pi(\bS^t) = \int \left[ \rE(\bS^t, \bA; \phi) + \gamma \VM^\pi(\bS^{t+1}) \big| \bS^{t+1} = f_{\text{trans}}(\bS^t, \bA; \phi) \right] \pi_{\theta}(\bA | \bS^t) \mathrm{d}\bA,
    \label{eqn:value}
\end{equation}
where the initial latent state is given as $\bS^t = f_{\text{enc}}(\bO^{1:t}; \phi)$.
%where we have used the fact that the transition dynamics is assumed to be deterministic, i.e.: $\bSp \triangleq \bS^{t+1} = f_{\text{trans}}(\bS, \bA)$. 
In the following we will describe our approach in two steps: We first describe how to learn the action-conditional model for $f_\text{enc}, f_\text{trans}$, $\rE$, and thus $\VM$ in Sec.~\ref{sec:model}. We then explain in Sec.~\ref{sec:policylearn} how the model can be used to optimize the policy. %We conclude with a discussion of the model's limitations (Section \ref{sec:model:limitations}).

\vspace{-0.2cm}
\subsection{Model Learning} \label{sec:model}
We need to estimate all quantities comprising Equation \eqref{eqn:value}; i.e., we need to estimate the following parametric functions (for brevity we use a single set of parameters $\phi$ for all model parameters):
\begin{center}
\begin{tabular}{ l l }
\textbf{Encode }$f_{enc}(\bO^{1:t}; \phi) = \bS^t \approx \mathbb{E}_{p(\sS | \bO^{1:t})}[\sS^t]$ &
\textbf{Transition }$f_{trans}(\bS^t, \bA; \phi) = \bS^{t+1}$ \\
\textbf{Decode }$f_{dec}(\bS^t; \phi) \approx \mathbb{E}_{p(\bO | \sS^t, \tau_{\leq t})}[\bO^t]$ &
\textbf{Value} $\VE^{\pi}(\bS^t; \phi) \approx V^\pi(\tau_{\leq t})$ \\
\multicolumn{2}{l}{\textbf{Reward} $\rE(\bS^t, \bA^t; \phi) = \rE(f_\text{trans}(\bS^t, \bA^t); \phi) \approx
\mathbb{E}_{p(\sS^t |  \tau_{\leq t}, \bA^t)}
[r(\sS^t, \bA^t)]$}
\end{tabular}
\end{center}
The \textbf{Encoder} maps a history of observations $\bO^{1:t}$ to a summary statistic or latent state $\bS^t$, via a recurrent neural network. 
%to learn the deterministic mapping into an \textit{abstract, unstructured latent representation}. 
Recurrence allows us to handle partially observable settings (eg: occlusion). 
The \textbf{Transition} function predicts the next latent state $\bS^{t+1}$ given the current state $\bS^t$ and action $\bA$, evolving dynamics in the low-dimensional latent space. %%We use a feed-forward fully-connected neural network as our transition model.
The \textbf{Decoder} maps the latent $\bS^t$ back to an expected observation $\bO^t$ and is primarily used as self-supervision for training the encoder \citep{wayne2018unsupervised}. %The decoder uses a de-convolutional architecture to reconstruct the images from latent state and fully-connected layers to predict the proprioception output.
The \textbf{Value} function, predicts the sum of expected rewards (the value) as a function of a latent $\bS^t$.
Lastly, the \textbf{Reward} function predicts the immediate, expected reward for a given latent state-action pair. Please refer to Section E of the supplementary for additional details on the model architecture. %Additional details on the architecture of the model components and a schematic of an $N$-step rollout through the model are provided in the supplementary. %(Sec.~\ref{sec:modelarch}).

%\item The \textbf{Policy} module learns the mapping from latent state to actions, predicting a distribution from which we can sample actions.
%To approximate Equation \eqref{eqn:value} -- and its gradient --  using these functions, we will have to define a loss based on trajectories acquired by interacting with the environment.
For the model-based value function $\VM$ in Eq.~\eqref{eqn:value} to form a good approximation of the true value $V^\pi$ (and its gradient) we train the model on trajectories collected while interacting with the environment.
%Assume, for now, that we have available a buffer $\mathcal{B}$ of such trajectory data, under an arbitrary behavior policy $\mu(\bA | \bS)$. 
The main approximation of our approach is the assumption that the evolution of the latent state is well modeled by a deterministic transition function. In partially observed and stochastic environments this is not guaranteed. For the relevant quantities to be well approximated despite this simplification we employ a number of losses that satisfy the following desiderata:
i) we want to ensure that $\bS^t$ is a sufficient statistic of the history $\bO^{1:t}$ and the system thus Markov in $\bS$; ii) we want to minimize the discrepancy between the predicted and observed evolution of the latent state $\bS$;
iii) given a latent state $\bS^t$ we want to accurately predict the expected reward and value (of policy $\pi$) at time step $t+k$ after executing some action sequence $\bA^{t:t+k}$.
%that we expect to obtain when executing policy $\pi$;
%iv) we want to minimize the discrepancy between our approximate value $\VE^\pi(\bS^t)$ and the true value $V^\pi(\sS^t)$.
%, this amounts to regressing rewards and values (for bootstrapping) from predicted latent states (c.f. $\mathcal{L}_r$, $\mathcal{L}_V$).
%The goal of learning is concisely described by two desiderata: 
%i) we want to ensure that latent state trajectories capture the expected dynamics of the true environment (see $L_f$ below) ii) we want to minimize the discrepancy between our approximate value $V^\pi(\bS^t)$ and the true value $V^\pi(\sS^t)$, this amounts to regressing rewards and values (for bootstrapping) from predicted latent states (c.f. $\mathcal{L}_r$, $\mathcal{L}_V$).
%

Let $\mathcal{B}$ denote a set of trajectory data collected while executing some behavior policy $\mu(\bA | \bS)$.
Let us define the full model loss after an initial ``burn-in'' of $H$ steps (to ensure the encoder has sufficient information) as $ \mathcal{L}^N = \mathbb{E}_{p_\pi,\mu} [ \sum_{t=H+1}^{H+N-1}  \mathcal{L}_e(\bS^t, \bA^t, r^{t}, \bO^{1:t+1}) |f_\text{trans}, f_{\text{enc}} ]$ where we approximate the expectation wrt. %$p_\pi(\sS^{t:T}, \bA^{t:T}, \bO^{t+1:T} | \tau_{\leq t})$ 
$p_\pi(r^{t:T}, \bA^{t:T}, \bO^{t+1:T} | \tau_{\leq t})$ 
with samples from $\mathcal{B}$,
\begin{equation}
%\begin{aligned}
    \mathcal{L^N} \approx \mathbb{E}_{\tau \sim \mathcal{B}} \left[ \sum_{t=H+1}^{H+N-1}  \mathcal{L}_e(\bS^t, \bA^t, r^{t}, \bO^{1:t+1}) \Big| \bS^{t+1} = f_\text{trans}(\bS^t, \bA^t; \phi), \bS^{H} = f_{\text{enc}}(\bO^{1:H}; \phi) \right],
    %\mathcal{L}^N = \mathbb{E}_{p_\pi,\mu} &\left[ \sum_{t=H+1}^{H+N} \mathcal{L}_f(\bS^t, \bO^{1:t}) + \alpha \mathcal{L}_{r}(\bS^t, 
    %\bA^t, r^t) + \beta \mathcal{L}_{V}(\bS^t, \bA^t, r^{t}, \bS^{t+1}) \Big| f_\text{trans}, f_{\text{enc}} \right] \\
    %\approx \mathbb{E}_{(\bO^{1:N}, \bA^{1:N}, r^{1:N}) \sim \mathcal{B}} &\left[
    %\sum_{t=H+1}^{H+N} \mathcal{L}_f(\bS^t, \bO^{1:t}) + \alpha \mathcal{L}_{r}(\bS^t, 
    %\bA^t, r^t) + \beta \mathcal{L}_{V}(\bS^t, \bA^t, r^{t}, \bS^{t+1}) \Big| f_\text{trans}, f_{\text{enc}} \right],
%\end{aligned} 
\label{eqn:modelloss}
\end{equation}
with $\tau := (\bO^{1:H+N}, \bA^{1:H+N}, r^{1:H+N})$ and where the per example loss is defined as $\mathcal{L}_e = \mathcal{L}_f(\bS^t, \bO^{1:t}) + \alpha \mathcal{L}_{r}(\bS^t, \bA^t, r^t) + \beta \mathcal{L}_{V}(\bS^t, \bA^t, r^{t}, \bO^{1:t+1})$. $\alpha$ and $\beta$ are %loss multipliers
coefficients that 
%defining 
determine the relative contribution of the loss components. The per example transition model loss is given as
\begin{equation}
\begin{aligned}
\mathcal{L}_f(\bS^t, \bO^{1:t}) = \| f_{\text{dec}}(\bS^t; \phi) - \bO^t \|_2^2 + \zeta \|f_{\text{enc}}(\bO^{1:t}; \phi) - \bS^t \|_2^2,
\end{aligned}
\end{equation}
where the first term measures the error between the observations $\bO$ and reconstructions from the open-loop latent state predictions ($\bS^{t>H}$) and the second term
enforces consistency between the latent state representation from the encoder $f_{\text{enc}}$ and the predictions from the transition model $f_\text{trans}$; this encourages the latent state to stay close to encodings of observed trajectories thus addressing points i) and ii) above. Here $\zeta$ is a coefficient that weights the two loss terms and the reward loss is 
\begin{equation}
\begin{aligned}
\mathcal{L}_{r}(\bS^t, \bA^t, r^t) = \| \rE(\bS^t, \bA^t; \phi) - r^t \|_2^2,
\end{aligned}
\end{equation}
where the value-loss is given by the, importance weighted, squared Bellman error 
\begin{equation}
\begin{aligned}
\mathcal{L}_{V}(\bS^t, \bA^t, r^{t}, \bO^{1:t+1}) = \frac{\pi(\bA^t | \bS^t)}{\mu(\bA^t | \bS^t)} \left(r^t + \gamma \VE^\pi(f_\text{enc}(\bO^{1:t+1}; \psi); \psi) - \VE^\pi(\bS^t; \phi)\right)^2,
\end{aligned}
\label{eqn:value_loss}
\end{equation}
where the next state value $\VE^\pi(f_\text{enc}(\bO^{1:t+1}; \psi); \psi)$ is calculated via a ``target network'', whose parameters $\psi$ are periodically copied from $\phi$, to stabilize training (see e.g. \cite{mnih2015human} for a discussion). In practice we use a v-trace target~\citep{espeholt2018impala}; see discussion in the supplementary. Note that both loss terms are evaluated for \emph{predicted} latent states with gradients flowing backwards through the transition model and eventually the encoder, which addresses point iii). We encourage the reader to consult to Section C \& Figure 1 in the supplementary material for additional details and a schematic.

% \begin{figure}
%     \centering
%     \resizebox{12cm}{!}{\input{figures/modelrollout.tex}}
%     \caption{\textit{Left}: Graph showing the rollout of a sequence of observations and actions $\tau$ (red circles), through the model (gray rectangles), predicting latent states and their corresponding reconstructed observations (blue circles). There are two stages: 1) Encoding the observations to the latent space (green block), followed by 2) an open loop rollout in the latent space, starting from the latent state $\bS^H$ (after a burn-in of length $H$) and applying the sequence of actions $\bA^{H:H+N-1}$ (yellow block). \textit{Right}: Actor execution of the policy. The past $H$ observations are encoded to generate state $\bS^t$ which is fed to the policy to generate action $\bA^t$.}
%     \label{fig:modelactorrollout}
% \end{figure}

% \subsection{Limitations}
% \label{sec:model:limitations}
% The main approximation in our modeling approach is the assumption that the evolution of the latent state is well modeled by a deterministic transition function. In partially observed and stochastic environments this is not guaranteed. However, the losses described in Section \ref{sec:model} are designed to ensure that the relevant quantities will be well approximated despite this simplification.

\vspace{-0.2cm}
\section{Imagined Value Gradients in Latent Spaces}
\vspace{-0.2cm}
\label{sec:policylearn}

\begin{figure}
    \centering
    \includegraphics[width=0.7\textwidth]{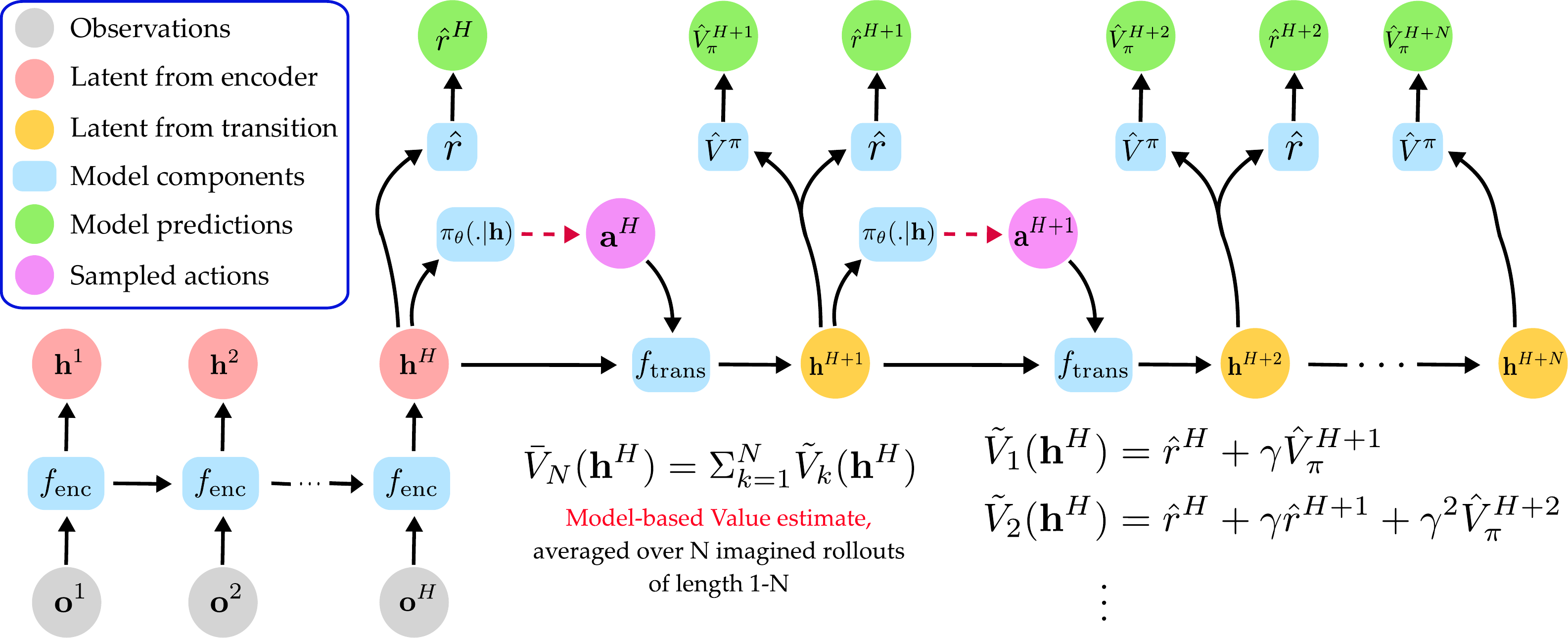}
    \caption{Imagined policy gradient computation. Given a history $H$ of observations from $\mathcal{B}$, we encode a latent state $\bS^H$, followed by an ``imagined'' rollout of length $N$ -- using sequence of sampled actions $\bA^{t>H}$ and $f_{trans}$. This leads to imagined states $\bS^{t>H}$ with corresponding value and reward estimates. We average cumulative rewards over $N$ horizons -- computing the estimate from Eq.~\eqref{eqn:svgn_avg_full}.}
    \label{fig:imaginedrollout}
    \vspace{-0.4cm}
\end{figure}

Given a model, we optimize a parametric policy $\pi_\theta(\bA | \bS)$ by maximizing the N-step surrogate value function which is a recursive composition of the policy, the transition, reward and value function:
\begin{equation}
    \VM_N(\bS^t) = \mathbb{E}_{\bA^k \sim \pi} \left[ \gamma^{N} \VE^\pi(\bS^{t+N}; \phi) + \sum_{k=t}^{t+N-1} \gamma^{k-t} \rE(\bS^k, \bA^k) \Big| \bS^{k+1} = f_\text{trans}(\bS^k, \bA^k) \right],
    \label{eqn:value_n}
\end{equation}
This N-step value can be computed by performing an ``imagined'' rollout in the latent state-space using our model (see ~\figref{fig:imaginedrollout}).
It can be maximized by gradient ascent, exploiting the so called ``value gradient''  $\nabla_\theta{\VM}_N(\bS^t)$ \citep{heess2015learning}; which can often be computed recursively, taking advantage of the reparameterization trick \citep{kingma2013auto,rezende2014stochastic} for sampling from $\pi_\theta$ and calculating analytic gradients via backpropagation backwards through time. We can express the policy as a deterministic function $\pi_\theta(\bS^t, \epsilon)$ that transforms a sample $\epsilon$ from a canonical noise distribution $p(\epsilon)$ into a sample from $\pi_\theta(\bA | \bS)$. In the following we will consider Gaussian policy distributions, i.e. $\pi_\theta(\bA | \bS) = \mathcal{N}(\mu_\theta(\bS), \sigma^2_\theta(\bS))$, for which the reparameterization is given as $\pi_\theta(\bS, \epsilon) = \mu_\theta(\bS) + \epsilon \sigma_\theta(\bS)$, with $p(\epsilon) = \mathcal{N}(\mathbf{0}, \mathbf{I})$, where $\mathbf{I}$ denotes the identity matrix. We can then calculate the gradient $\nabla_\theta\VM_N(\bS^t)$ for any state as
% \begin{equation}
%  \nabla_\theta{\VM}_{N}(\bS^t) = \! \! \!\!\! \sum_{k=t}^{t+N-1} \!\!\mathbb{E}_{p(\epsilon)} \Big[  \nabla_\theta \rE(\bS^{k}, \pi_\theta\big(\bS^{k}, \epsilon)\big) + \gamma \nabla_{\bS^{k+1}} {\VM}_{N-k}(\bS^{k+1}) \nabla_\theta f_\text{trans}\big(\bS^{k}, \pi_\theta(\bS^{k}, \epsilon)\big) \Big],
% \label{eqn:svgn_theta}
% \end{equation}
% \begin{equation}
% \nabla_\theta{\VM}_N(\bS^t) = \mathbb{E}_{p(\epsilon)} \big[\nabla_\theta \rE\big(\bS^{t}, \pi_\theta(\bS^{t}, \epsilon)\big) + \gamma \nabla_{\bS^{t+1}} \VM_{N-1}(\bS^{t+1}) \nabla_\theta f_\text{trans}\big(\bS^{t}, \pi_\theta(\bS^{t}, \epsilon)\big) + \gamma \nabla_\theta{\VM}_{N-1}(\bS^{t+1}) \big]
% \label{eqn:svgn_theta}
% \end{equation}
%\begin{equation}
%\nabla_\theta{\VM}_N(\bS^t) = \mathbb{E}_{p(\epsilon)} \big[\nabla_\theta \rE\big(\bS^{t}, \pi_\theta(\bS^{t}, \epsilon)\big) + \gamma \nabla_{\bS^\prime} \VM_{N-1}(\bS^\prime) \nabla_\theta f_\text{trans}\big(\bS^{t}, \pi_\theta(\bS^{t}, \epsilon)\big) + \gamma \nabla_\theta{\VM}_{N-1}(\bS^\prime) \big]
%\label{eqn:svgn_theta}
%\end{equation}
\begin{equation}
\nabla_\theta{\VM}_N(\bS^t)\!\! = \!\!\mathbb{E}_{p(\epsilon)} \big[\big(\nabla_\bA \rE\big(\bS^{t}, \bA\big)\! +\! \gamma \nabla_{\bS^\prime} \VM_{N-1}(\bS^\prime) \nabla_\bA f_\text{trans}\big(\bS^{t}, \bA\big)\big) \nabla_\theta \pi_\theta(\bS^{t}, \epsilon)\! + \! \gamma \nabla_\theta{\VM}_{N-1}(\bS^\prime) \big]
\label{eqn:svgn_theta}
\end{equation}
where $\bS^\prime := \bS^{t+1} = f_\text{trans}(\bS^{t}, \pi_\theta(\bS^{t}, \epsilon))$ and we dropped the dependencies on $\phi$ for brevity. The value gradient $\nabla_{\bS^\prime} {\VM}_{N-1}(\bS^\prime)$ wrt. a state $\bS^\prime$ is defined recursively and provided in Eq.~(3) in the supplementary material.
The case $\nabla_\theta{\VM}_{1}(\bS^{t+N})$ is established by assuming the policy is fixed in all steps after $N$; i.e. bootstrapping with $\nabla_{\bS^{t}}\VM_0(\bS^{t}) = \nabla_{\bS^{t}}\VE^\pi(\bS^{t}; \phi)$. To calculate the gradient, only an initial state $\bS^t$ (encoded from $\bO^{1:t} \sim \mathcal{B}$) is required (see \figref{fig:imaginedrollout}). Eq.\ \eqref{eqn:svgn_theta} computes $N$ policy gradient contributions for the encoded state $\bS^t$ as well as for the imagined states $\bS^{t+1} \dots \bS^{t+N-1}$. This ensures that the policy can be evaluated on either kind of latent state. Our derivation is analogous to SVG \citep{heess2015learning}; but using imagined latent states and assuming a deterministic transition model.

%Talk about policy, value \& reward functions. Then given a sequence of observations, actions \& rewards we discuss how to train the entire system together (including the model).

%Model uses the rollout and losses from above. Reward uses the true reward basically (from trans model states?). Value uses target networks + retrace but we predict values from model rollouts and compare it to target network predictions on true encoded states or the model rollouts. Which? Write out whole loss and mention that model is trained jointly and all losses here backprop to model so that representation is good for predicting value \& reward. We can discuss fitted value learning etc in supplementary.

%Using sequence of observations, actions and rewards. Show how the rollout works, write out the math for the value function learning, write out the losses for the model \& reward network, mention that these parameters share gradients so model gets gradients from reward, value, decoder etc. Add a fitted value learning algorithm similar to the SVG paper in supplementary. Mention that thsi part is pretty much same as SVG but with Retrace \& latent representation learning. Talk about burn-in / history with the recurrent model.

\vspace{-0.2cm}
\subsection{Stable Regularized Policy Optimization}
We make the following observations regarding the use of value gradients in practice:
%We make a few interesting observations relevant to the value gradients use in practice. 
First, we can obtain different gradient estimators by varying $N$. For small $N$ we obtain a more biased value gradient -- due to the reliance on bootstrapping -- that changes slowly (i.e. it changes at the speed of convergence for $\VE^\pi(\bS; \phi)$). For large $N$, less bias, and faster learning could be achieved if model and reward predictors are accurate, but the estimate can be affected by modelling errors. As a compromise, we found that averaging gradient estimates obtained with different-length rollouts worked well in practice
$
    \nabla_\theta{\bar{V}}_N(\bS) = \frac{1}{N} \sum_{k}^N \nabla_\theta{\VM}_k(\bS).
    \label{eqn:svgn_avg}
$
We refer to the supplementary for details.

Second, even with this averaged gradient, optimization is prone to exploiting modelling errors (in the transition dynamics and reward/value estimates) -- see e.g. \citep{planet-dm} for a discussion. To counteract this, we employ relative-entropy (KL) regularization. Similar to existing policy optimization methods \citep{schulman2017proximal,abdolmaleki2018maximum,haarnoja2018soft} we augment the estimated reward with a KL penalty, yielding
$
    \rE_\text{KL}(\bS, \bA, \pi_\theta) = \rE(\bS, \bA) + \lambda \log\frac{\pi_\theta(\bA | \bS)}{p(\bA | \bS)},
$
where $p(\bA | \bS)$ is a the prior action probability (we use $p(\bA | \bS) = \mathcal{N}(\mathbf{0}, \mathbf{I})$ throughout) and $\lambda$ is a cost multiplier. Replacing $\rE$ in Equation \eqref{eqn:svgn_theta} with $\rE_\text{KL}$ -- noting that $\rE_\text{KL}$ is differentiable wrt. $\theta$ -- results in the regularized value gradient $\nabla_\theta{\VM}^{\text{KL}}_{N}(\bS)$.
Analogously, we obtain a compatible value ${\VM}^{\text{KL}}_{0}(\bS)$ by replacing $r$ with $r_\text{KL}$ in Equation \eqref{eqn:value_loss}. The total derivative estimate we use then is
\begin{equation}
\textstyle
       \nabla_\theta \mathbb{E}_{p_{\pi}} \Big[ \bar{V}^{\text{KL}}_N(\bS^t) \Big] \approx \mathbb{E}_{\bO^{1:t} \sim \mathcal{B}} \Big[ \frac{1}{N} \sum_{k=1}^N \nabla_\theta{\VM}^\text{KL}_k(f_\text{enc}(\bO^{1:t})) \Big],
    \label{eqn:svgn_avg_full}
\end{equation}
where we use batches from the replay to optimize the policy on all visited states -- using Eq.~\eqref{eqn:svgn_avg_full} in combination with Adam \cite{kingma2014adam}. 
%The full learning procedure involves executing the most recent policy to populate the replay buffer $\mathcal{B}$ in parallel with model and policy optimization using the model loss from Eq.~\eqref{eqn:modelloss} and the averaged policy gradient from Eq.~\eqref{eqn:svgn_avg_full}. 
Please refer to Algorithm (1) in the supplementary for a  description.

\begin{figure}
\centering
  \includegraphics[width=.9\linewidth]{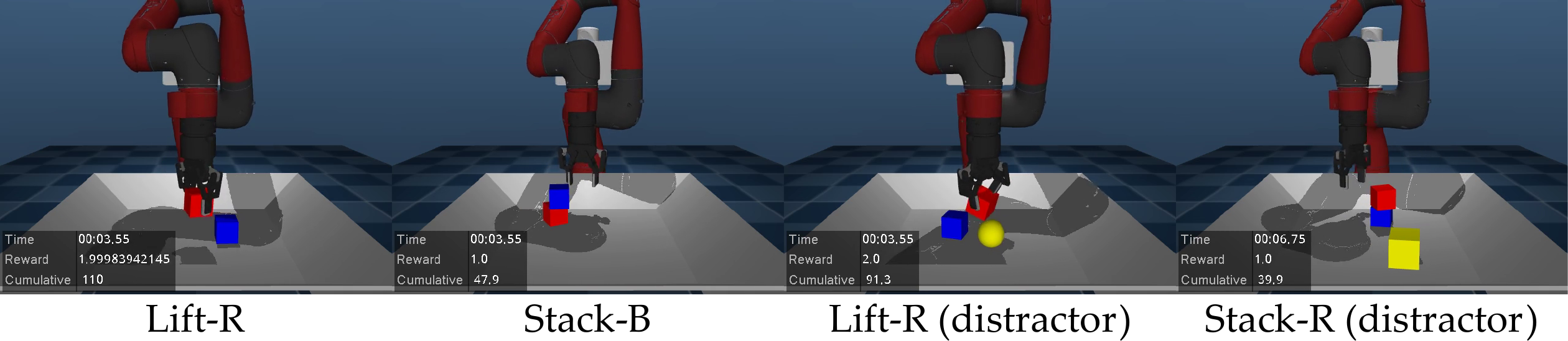}
    \caption{\textit{Left}: \textbf{Lift-R} task - robot lifts the red block. \textit{Center-Left}: Example scene from the \textbf{Stack-B} task. \textit{Center-Right} and \textit{Right}: Tasks with unseen \textbf{distractors}. %(yellow sphere and yellow cube) added to the scene for the Lift-R and Stack-R tasks.
    }
    \label{fig:task_setup}
    \vspace{-2mm}
\end{figure}

\vspace{-0.2cm}
\section{Experiments} \label{sec:results}
\vspace{-0.2cm}
We evaluate our approach on several challenging long-horizon manipulation tasks in simulation (see sections D \& F of the supplementary for details). %in the Mujoco simulator \cite{todorov2012mujoco}. 
Tasks involve the agent controlling a Sawyer manipulator equipped with a Robotiq gripper with a 5-dim.\ control ($N_A $= 5) to interact with a red and blue block on a tabletop. Observations are 64x64px RGB images from two cameras located on either side of the table, looking at the robot,
and proprioceptive features.
The latent representation is 128-dimensional ($N_H$=128) and unless otherwise noted we use a history of $H=3$ observations and a rollout horizon of $N=5$.

\textbf{Task Setup: } \figref{fig:task_setup} presents visualizations of a subset of tasks from our experiments. We consider three main tasks: 1) the \textbf{Lift} task requires the robot to lift an object above a certain threshold; \textit{Lift-R} refers to lifting the red block and analagously for \textit{Lift-B}. 2) the \textbf{Stack} task requires the robot to stack one object on top of the other; \textit{Stack-R} refers to stacking the red block on top of the blue block and vice-versa for \textit{Stack-B}. 3) Lastly, the \textbf{Match Positions} task involves moving both objects to a fixed target position. We also consider variants of these tasks with the addition of \textbf{visual distractors} and \textbf{stochasticity}. It is worth noting that all our tasks involve long-term dependencies and complicated contact dynamics making them particularly challenging for model-based approaches. We use shaped rewards for all tasks except the \textit{Match Positions} task which has a mixed dense-sparse reward.

\textbf{Baselines: } We consider the following pixel-based baselines for our experiments: 1) \textbf{SVG(0)}: the model-free version of SVG. As our approach (termed Imagined Value Gradients -- \textbf{IVG} from here on out) builds on SVG we expect to improve on SVG(0). 2) \textbf{MPO}: Maximum a Posteriori Policy Optimisation \cite{abdolmaleki2018maximum}, a state-of-the-art model-free approach. To obtain an upper bound on performance we also include a version of MPO with access to the full system state (incl.\ objects). For the transfer experiments we also experiment with variants where we replace the value gradient based optimization. In particular we use \textbf{CEM}: the cross-entropy method \citep{Rubinstein:2004:CEM:1014902} using the same model as IVG for transfer (latent rollouts). \textbf{PG}: replacing the value gradients with a likelihood ratio estimator (using 100 imagined rollouts), again using the same model as used for the \emph{IVG} transfer. Additional details on the baselines are given in the appendix.

\vspace{-0.2cm}
\subsection{Learning from scratch}
\begin{figure}
\begin{subfigure}{.33\textwidth}
  \centering
  \includegraphics[width=.95\linewidth]{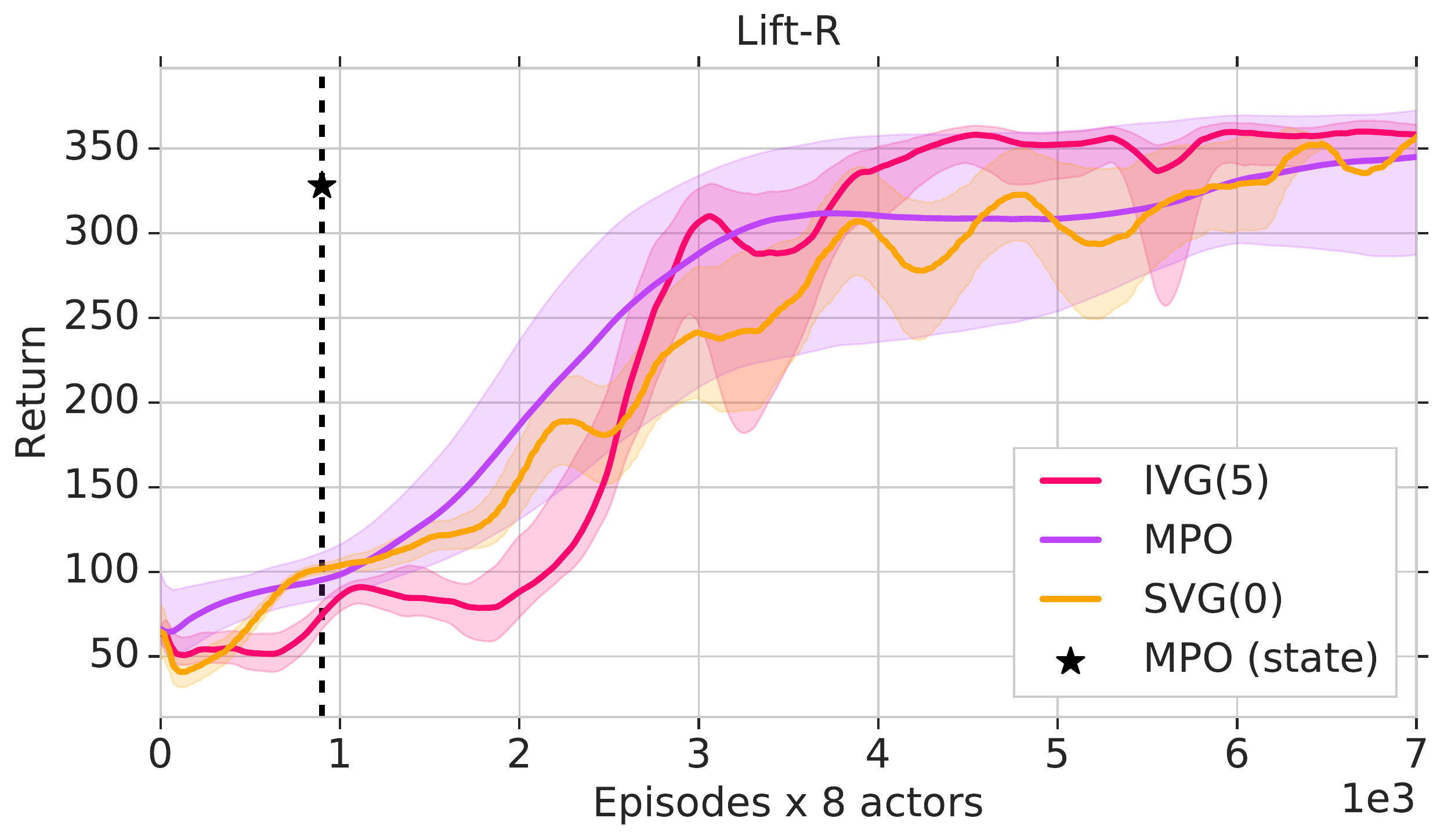}
\end{subfigure}%
\begin{subfigure}{.33\textwidth}
  \centering
  \includegraphics[width=.95\linewidth]{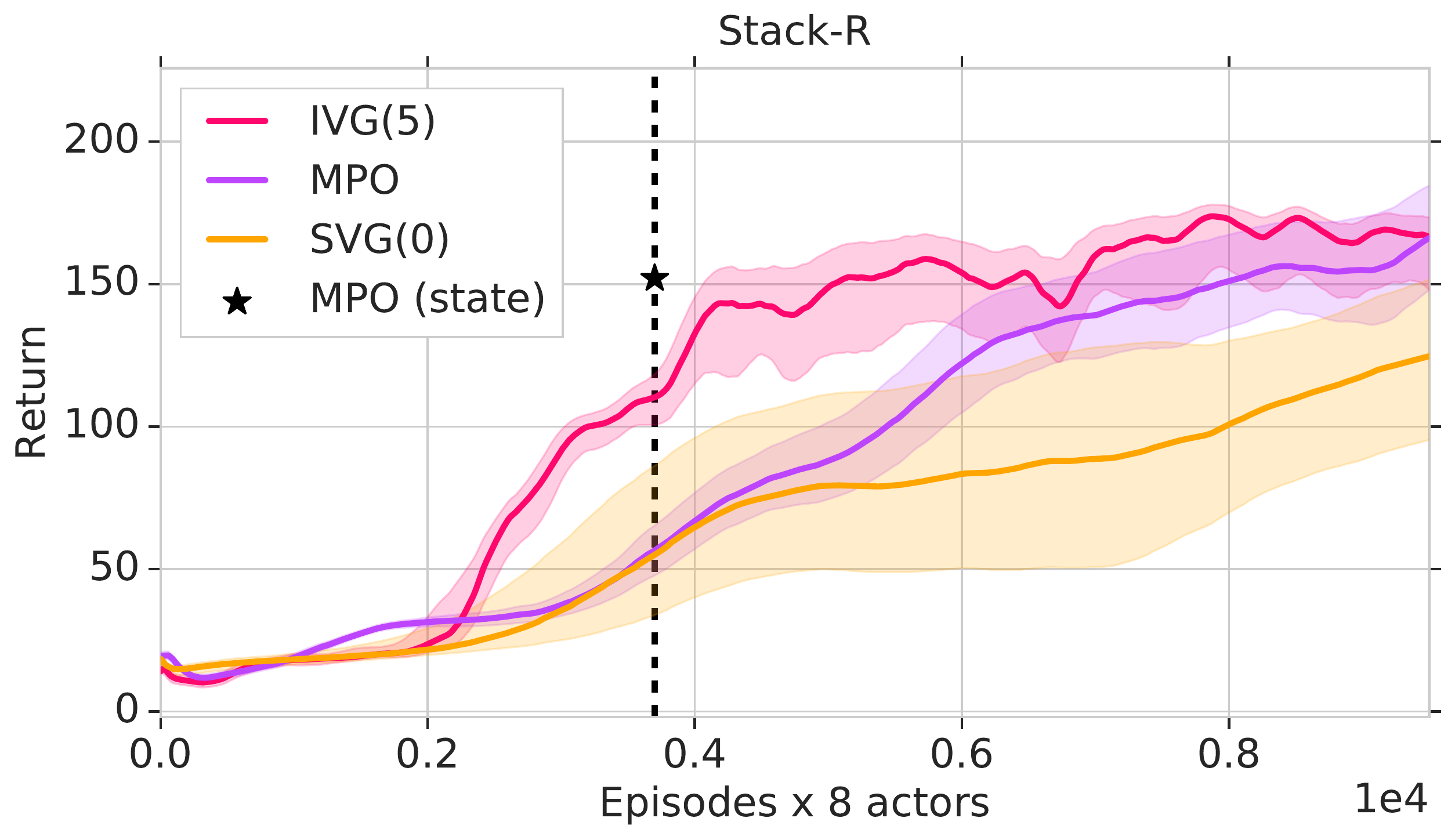}
\end{subfigure}
\begin{subfigure}{.33\textwidth}
  \centering
  \includegraphics[width=.95\linewidth]{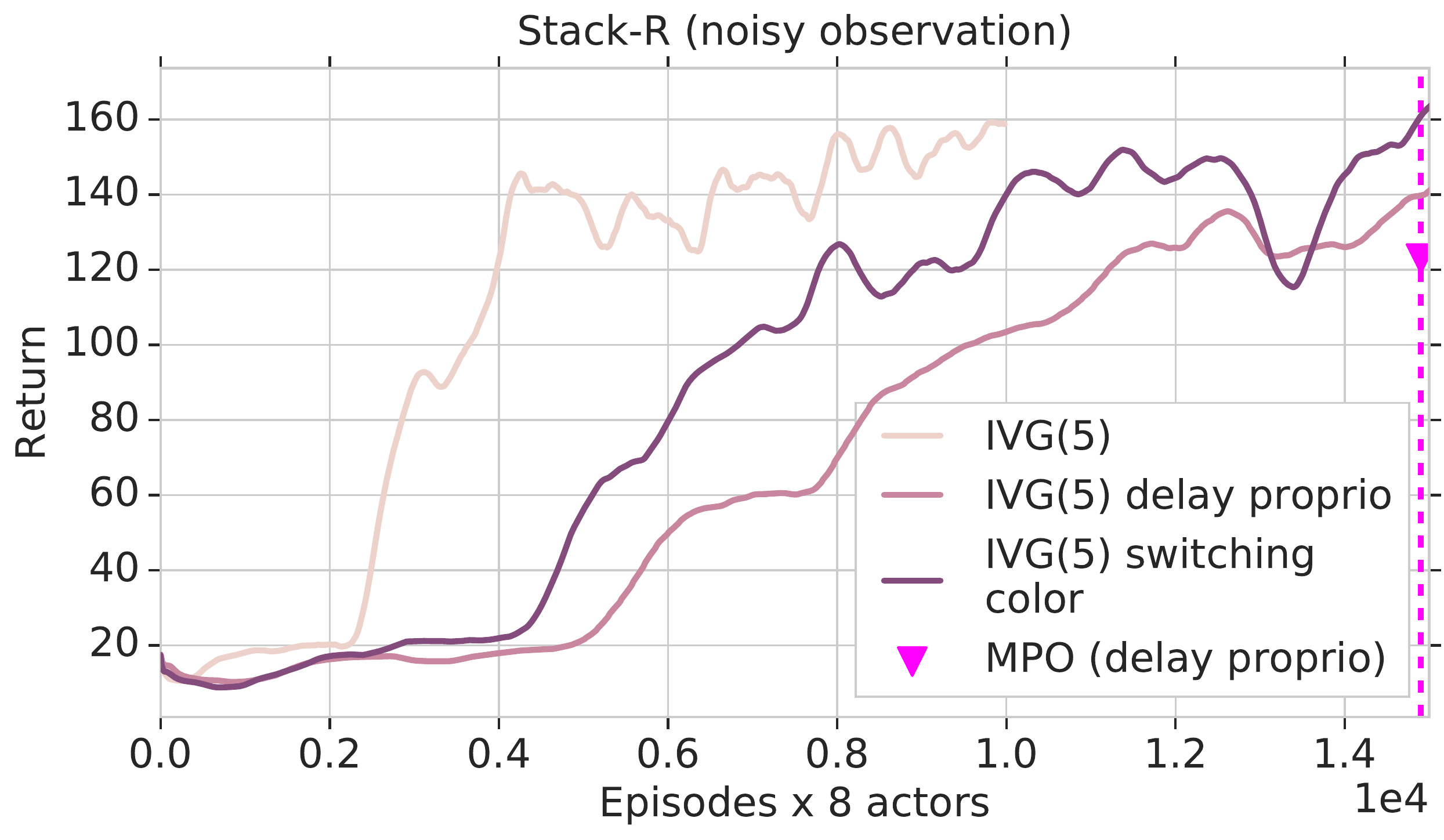}
\end{subfigure}
\caption{\textit{Left}: IVG(5), with $N=5$, performs similarly to the model-free baselines MPO and SVG(0) on the \textbf{lift} task. Black dotted line represents MPO (state) scoring higher than a reward threshold indicating success (300 for Lift, 140 Stack) for 50 episodes. \textit{Center}: IVG(5) learns significantly faster than SVG(0) and MPO on the \textbf{stack} task. \textit{Right}: IVG(5) learns well even in environments with noisy observations.}
\label{fig:scratch_results}
\vspace{-0.5cm}
\end{figure}

We first compare IVG and the baselines when learning the Lift-R and Stack-R manipulation tasks \textit{from scratch}:
Model-based IVG(5) learns the simpler \emph{lift} task stably and performs on par with the baselines (\figref{fig:scratch_results}, left). 
On the harder stack task (\figref{fig:scratch_results}, right), IVG(5) learns significantly (about 2x) faster than both MPO and SVG(0), and also outperforms SVG(0) in terms of final performance.
Compared to the informed MPO (State) baseline, IVG learns more slowly, but this difference is significantly reduced for stacking. Even when learning from pixels, the structure inherent in IVG via the latent space rollouts allows it to learn complex tasks faster than strong model-free baselines. 

We also tested the ability  of IVG to handle slightly more stochastic and partially observed environments.
\figref{fig:scratch_results} (right) presents results on learning the Stack-R task in environments with: 1) delayed proprioception (2 timestep lag) and 2) noisy observations where one of the blocks switches colors randomly every 3 frames. IVG(5) successfully learns in both cases (albeit more slowly). MPO also solved these tasks but required approximately $2x$ more episodes than IVG (color switch not shown).

\vspace{-0.2cm}
\subsection{Transferring learned models to related tasks}
IVG learns a model of the environment which we may be able to transfer, and thus accelerate learning of related tasks. In the following, we evaluate this possibility.
First, we train IVG from scratch on one or multiple source task(s). 
Second, we copy the weights of the trained model (\textbf{encoder} $f_\text{enc}$, \textbf{transition} model $f_\text{trans}$ and \textbf{decoder} $f_\text{dec}$). The policy, value and reward models are not transferred and are learned from scratch.
%We then use the previously copied weights to initialize the model parameters when training on the target task, whose performance is reported in the experiments. 
As a model-free baseline we include MPO where we initialize all weights of the policy \& value function except the last layer from an agent trained on a source task.
%A major advantage of IVG, in comparison to model-free methods, is that we learn a model of the environment; this model can be used to accelerate learning on related tasks. In the following, we evaluate the transfer capability of models learned using IVG. Briefly, transfer happens in three steps: 1) First, we train IVG from scratch on a source task (or multiple source tasks). 2) Second, we copy the weights of the trained model, specifically of the \textbf{encoder} $f_\text{enc}$, \textbf{transition} model $f_\text{trans}$ and \textbf{decoder} $f_\text{dec}$. The policy, value and reward models are not transferred and are learned from scratch. 3) We use the previously copied weights to initialize the model parameters when training on the target task, whose performance is reported in the experiments. As an additional baseline, we train a version of MPO where we initialize all weights of the policy \& value function (except the last layer) from an agent trained on a prior source task.

\textbf{Multiple source tasks: } A model trained on multiple tasks should transfer better due to the diversity in transitions observed. To test this hypothesis, we propose a version of IVG that is trained on multiple source tasks -- learning a task agnostic model. We use the following source tasks: Reach-B, Move-B, Lift-B, Stack-B and Reach-R (most tasks involve the blue object but Reach-R gives  the
model some experience with the red block). Details on this setup are provided in the supplementary.

\begin{figure}
\begin{subfigure}{.33\textwidth}
  \centering
  \includegraphics[width=.95\linewidth]{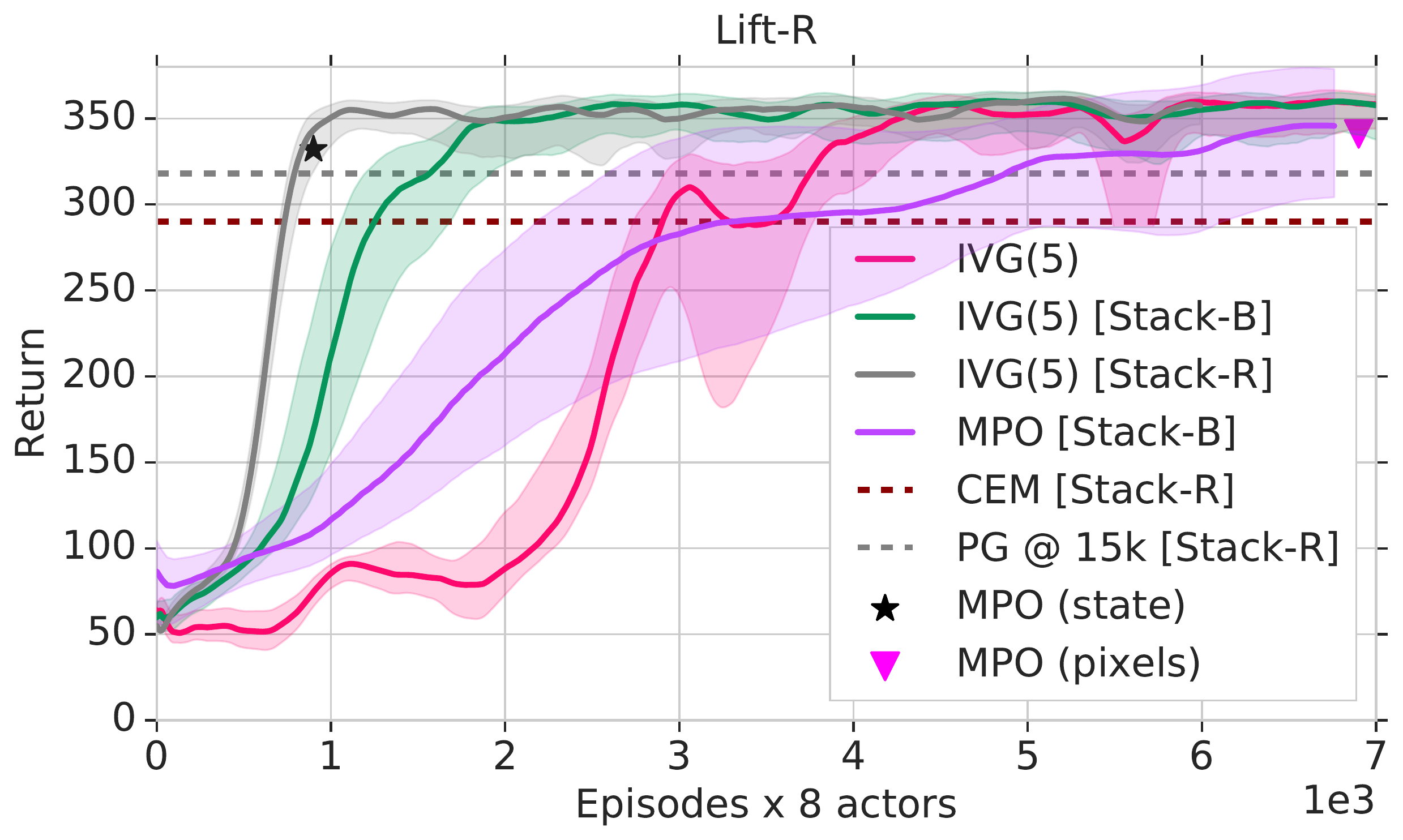}
\end{subfigure}%
\begin{subfigure}{.33\textwidth}
  \centering
  \includegraphics[width=.95\linewidth]{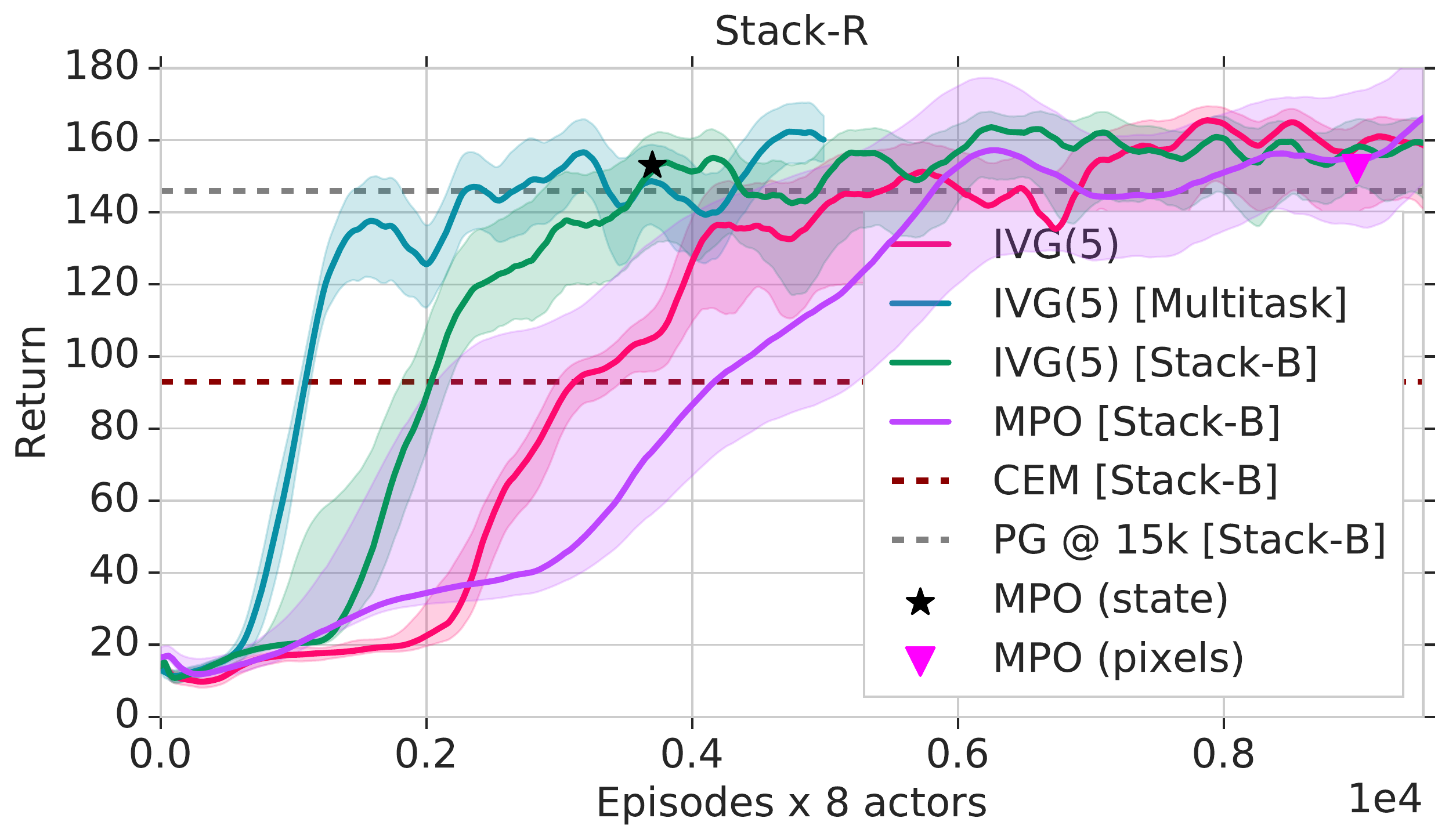}
\end{subfigure}%
\begin{subfigure}{.33\textwidth}
  \centering
  \includegraphics[width=.95\linewidth]{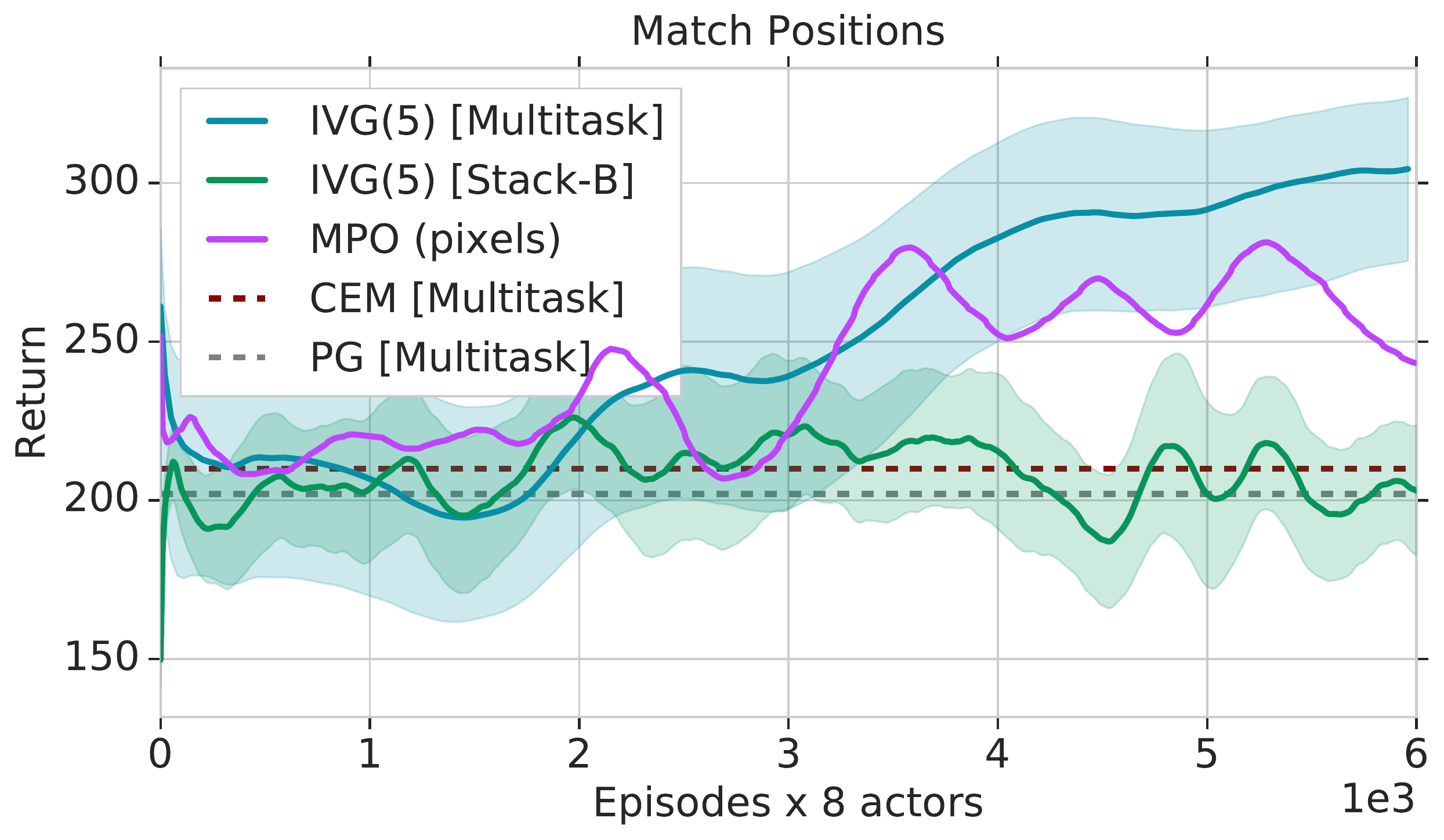}
\end{subfigure}
\caption{Transfer performance of IVG(5) on various target tasks. Task names inside square brackets indicate the source task. \textit{Left}: Transfer results on Lift-R. Transferring from Stack-B leads to large improvements compared to baselines. \textit{Center}: Transfer from multi-task IVG outperforms all baselines and single-task IVG. \textit{Right}: Multi-task IVG successfully transfers on the Match Positions task while single-task IVG and other baselines fail.}
\label{fig:transfer_results}
\vspace{-2mm}
\end{figure}

\textbf{Transfer results: } We present results on transferring IVG models to the following target tasks: 

%1) \textbf{Lift-R: } \figref{fig:transfer_results} (left) shows the transfer performance of IVG(5) on the Lift-R task. When transferring from the source task \textit{Stack-B}, IVG(5) learns $\sim$2x faster compared to learning from scratch and $\sim$4x faster than MPO (irrespective of whether it is pre-trained or not). Furthermore, transferring IVG(5) from the source task \textit{Stack-R} accelerates learning further as the model has already observed many relevant transitions, achieving speed comparable to the state-based version of MPO. Both these results highlight the significant benefits of a model when transferring to related tasks, even in cases where the model has only observed a subset of relevant transitions. 
1) \textbf{Lift-R: } \figref{fig:transfer_results} (left) shows the transfer performance of IVG(5) on the Lift-R task. With a model pre-learned on \textit{Stack-B}, IVG(5) learns $\sim$2x faster than from scratch and $\sim$4x faster than MPO (irrespective of MPO's initialization). A model pre-trained on \textit{Stack-R} accelerates IVG(5) further since the model has already observed many relevant transitions, achieving speed comparable to \emph{MPO (state)}. Replacing the learned policy with direct optimization (\emph{CEM}) or using a policy gradient yields sub-optimal behavior. These results highlight the benefits of a model when transferring to related tasks, even in cases where the model has only observed a subset of relevant transitions.

%2) \textbf{Stack-R: } Similar to the results on the Lift-R task, IVG with transferred models learns much faster on the Stack-R task (\figref{fig:transfer_results}, center). In addition to transfer from \textit{Stack-B}, we also compared the performance when transferring from a model trained on multiple source tasks (\textit{Multitask}). Crucially, this multi-task variant significantly accelerates learning speed; it is $\sim$1.5x faster than transferring from a single-task and about $\sim3x$ faster than learning from scratch. As we will see from subsequent results, models trained on multiple tasks greatly accelerate transfer.
2) \textbf{Stack-R: } Results with transferred models for Stack-R are similar to those for Lift-R (\figref{fig:transfer_results}, center). But here, \emph{CEM} fails to perform the stack (possibly caused by overly exploiting the model due to missing policy regularization), and using \emph{PG} instead of a value gradient takes significantly longer to converge (15k trajectories) and performs worse, likely due to the noise in the likelihood ratio calculation -- even though we already used 100 forward rollouts for the likelihood ratio calculation (1 for IVG), a $25$ fold increase in computation. In addition, we also compared the performance when transferring a model trained on multiple source tasks (\textit{Multitask}). This multi-task variant significantly accelerates learning speed; it is $\sim$1.5x faster than transferring from a single-task and about $\sim3x$ faster than learning from scratch. As we will see from subsequent results, models trained on multiple tasks greatly accelerate transfer.

3) \textbf{Match Positions: } We tested model generalization on the \emph{match positions} task, which differs significantly from the source tasks (\figref{fig:transfer_results}, right). All agents except  multi-task IVG fail on this task; this includes transferring IVG from a single task (Stack-B) and the pixel-based MPO. This is likely due to the sparse structure of the reward (see supplementary). This shows how multi-task training enables the formation of robust and expressive latent spaces that transfers well to new tasks.

%3) \textbf{Match Positions: } We tested the generalization of our learned models on the match positions task, a task with different reward structure than all prior tested tasks (\figref{fig:transfer_results}, right). All agents except the one transferring from multi-task IVG fail on this task; this includes transferring IVG from a single task (Stack-B) and the pixel-based MPO. We believe this is due to the sparseness of the reward; significant exploration is needed to figure out the reward substructure.  This further demonstrates the benefits of multi-task learning; representations learned on multiple tasks need to be predictive of diverse rewards and values, this enables robust and expressive latent space learning.

%\textbf{Visual distractors: } Lastly, we analyzed the generalization of IVG in the presence of visual distractors; \figref{fig:distractor_ablation}, (left \& center) presents results when a yellow cube or ball are added to the scene. IVG, when transferring from single or multiple source tasks is still able to learn stably on these tasks (as does IVG from scratch). The multi-task transfer version of IVG significantly outperforms all other methods, learning $\sim$3x faster than from scratch and $\sim$4x faster than MPO. This is a surprising result as CNNs are known to be sensitive to changes in visual inputs; jointly training a model with the policy allows it to optimize for task performance in addition to prediction and reconstruction, leading to more robust representations that can transfer quickly. 

\begin{figure}
\begin{subfigure}{.33\textwidth}
  \centering
  \includegraphics[width=.95\linewidth]{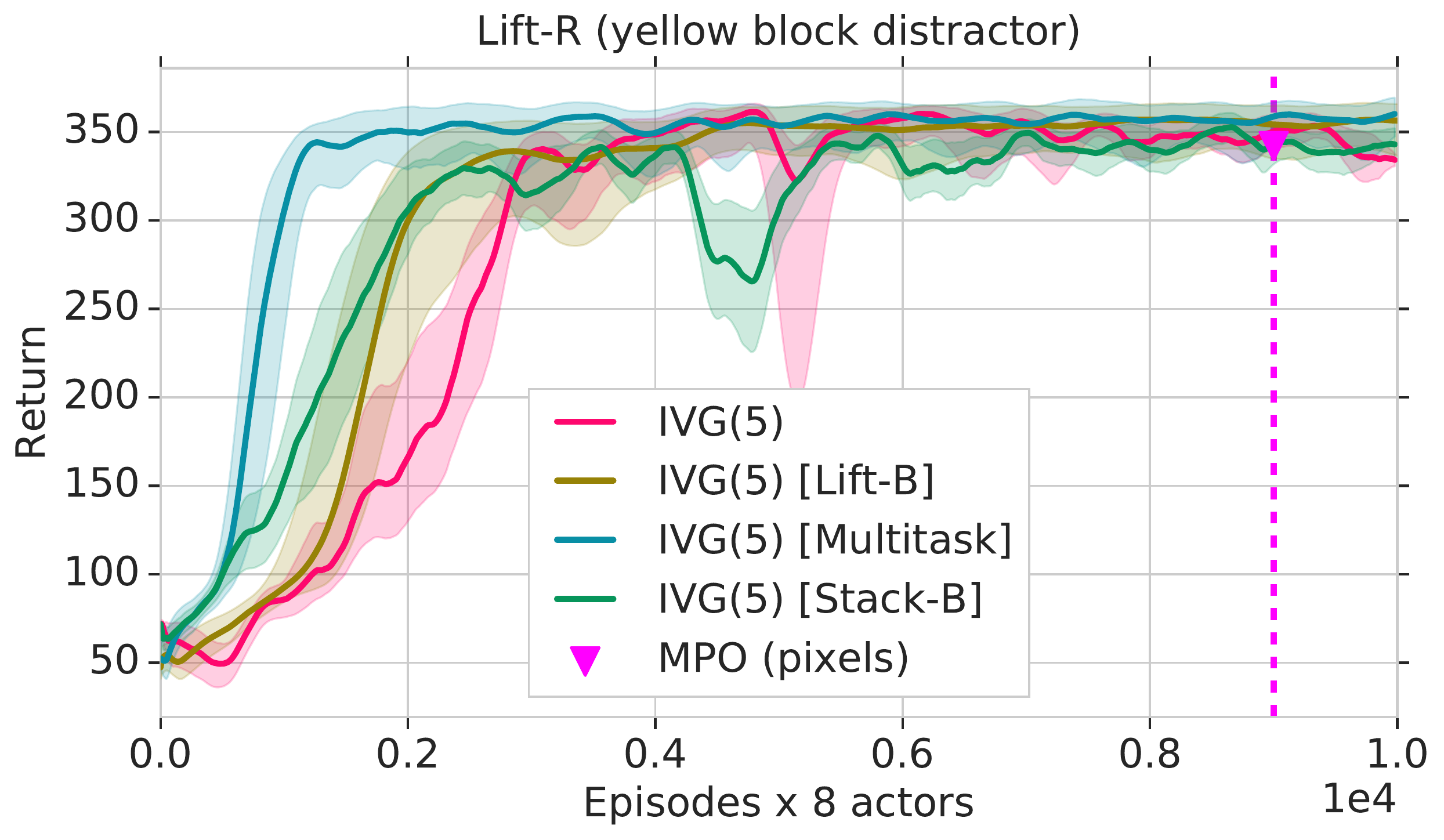}
\end{subfigure}%
\begin{subfigure}{.33\textwidth}
  \centering
  \includegraphics[width=.95\linewidth]{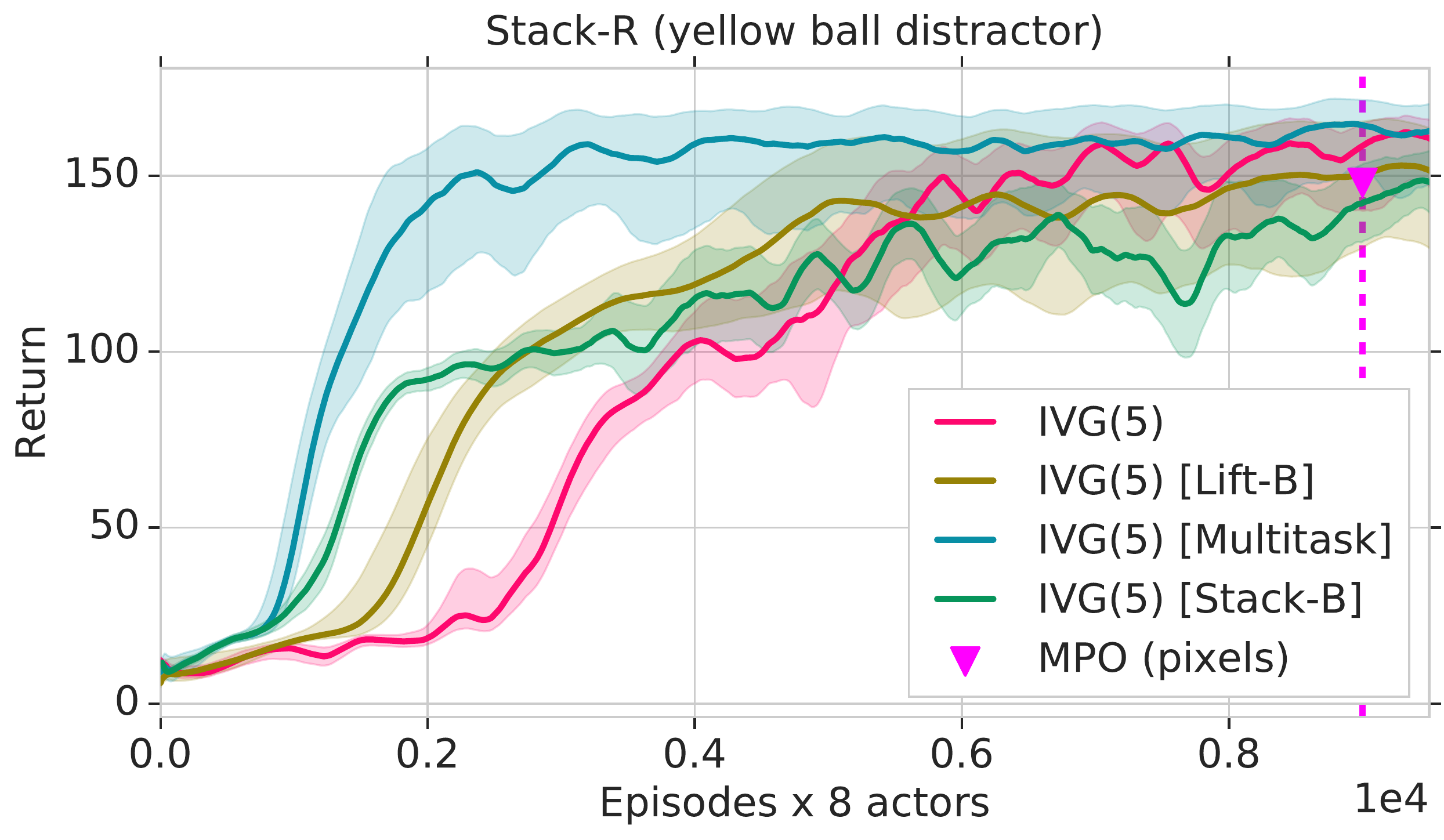}
\end{subfigure}%
\begin{subfigure}{.33\textwidth}
  \centering
  \includegraphics[width=.95\linewidth]{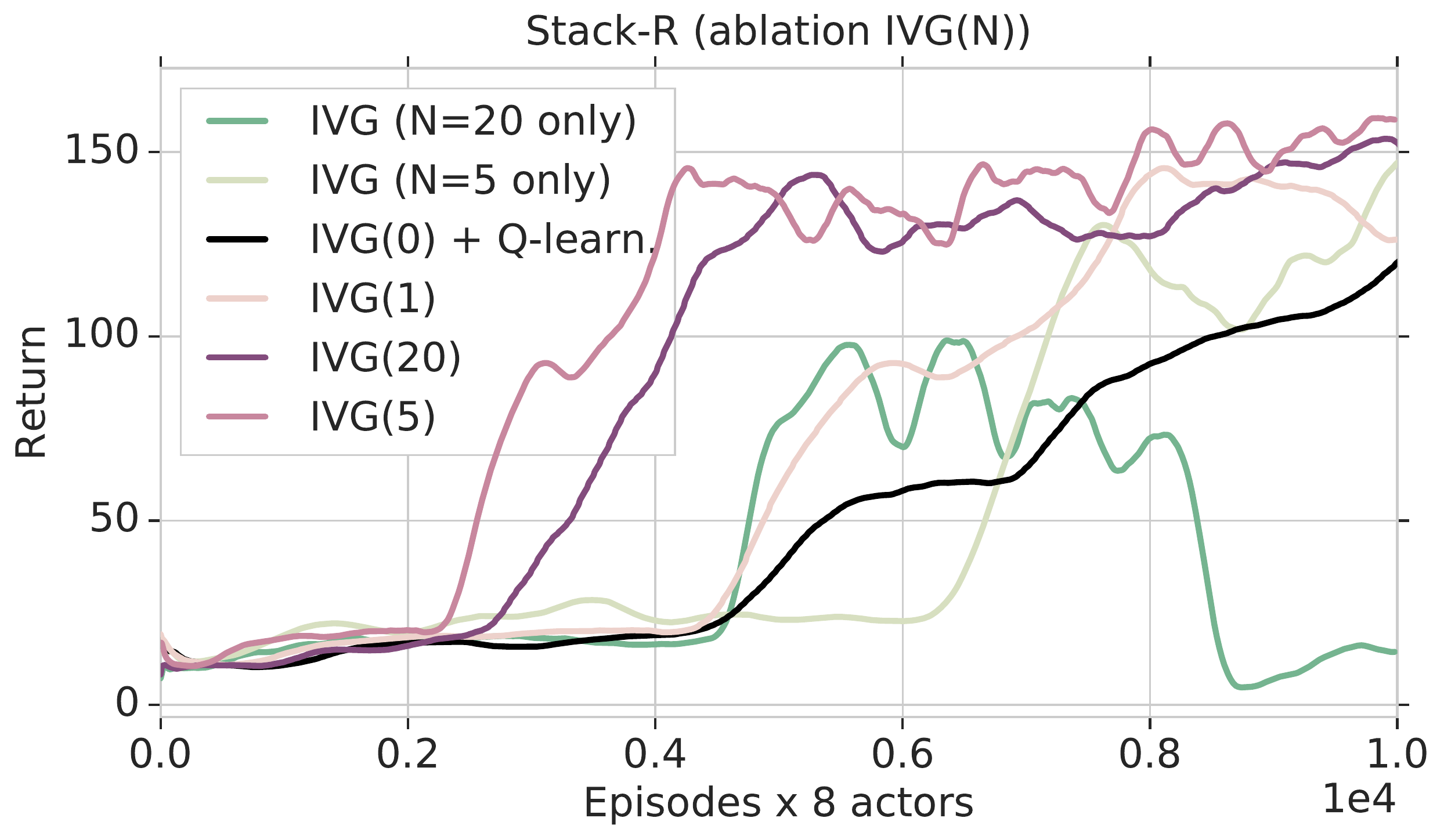}
\end{subfigure}
\caption{\textit{Left \& Center}: Transfer performance on the visual distractor task where a yellow block or ball are added to the Lift-R and Stack-R task respectively. 
transfer from multi-task IVG leads to a $\sim$3-4x speedup \textit{Right}: Ablation tests with IVG, using horizon lengths of $N=0,1,5,20$ \& using a single N-step horizon for computing the policy gradient (Eq.~\eqref{eqn:svgn_theta} instead of Eq.~\eqref{eqn:svgn_avg_full}). 
IVG(0) uses the model loss (with horizon 5) only as an auxiliary signal (and otherwise corresponds to SVG(0)). 
}
\label{fig:distractor_ablation}
\vspace{-2mm}
\end{figure}

4) \textbf{Visual distractors: } Lastly, we analyzed the generalization of IVG when visual distractors in the form of a yellow cube or ball are added to the scene (\figref{fig:distractor_ablation}, (left \& center)). Transfer from one or multiple source tasks remains effective. Transferring a multi-task model significantly outperforms all other methods, learning $\sim$3x faster than from scratch and $\sim$4x faster than MPO. Thus, even though CNNs are known to be sensitive to changes in visual inputs, jointly training a predictive model with the policy  can still lead to robust representations that transfer quickly.

\vspace{-0.2cm}
\subsection{Ablation experiments}
\label{sec:ablation}
%\akb{Note: corresponding plots still missing}
% We further analyzed the performance of IVG with two ablations, evaluating our design choices:

% \textbf{Rollout horizon: } To evaluate the effect of the rollout horizon $N$, we tested IVG across multiple settings of $N$=0,1,5,20. \figref{fig:distractor_ablation} (right) presents these results, evaluated on learning the Stack-R task from scratch; in general IVG is robust to the choice of the rollout horizon, i.e. IVG learns stably for all settings of $N$ we tried. Interestingly, the choice of $N$ has a marked effect on the speed of learning, smaller values $N$=0,1 tend to learn slowly; increasing $N$ speeds up learning dramatically (see $N$=5) upto a point after which there is a potential slowdown. 

% \textbf{Averaging value gradients: } To quantify the effect of averaging value gradients across multiple rollout horizons \eqref{eqn:svgn_avg_full}, we ran experiments using a single imagined rollout of length $N$. \figref{fig:distractor_ablation} (right) presents these results; for short horizons e.g. $N$=5, using a single rollout horizon leads to lower learning. On the other hand, learning completely fails for longer horizons $N$=20 (unlike with averaging) potentially due to cascading model errors on imagined rollouts, validating our approach.

To validate our algorithm and model design choices we performed two sets of ablations.\\
\textbf{Rollout horizon: } To evaluate the effect of the rollout horizon $N$ on the policy gradient (eq.~ \eqref{eqn:svgn_theta}) we tested IVG across multiple settings of $N$=0,1,5,20 for learning \emph{Stack-R} from scratch ( \figref{fig:distractor_ablation} (right)). IVG is robust to the choice of the rollout horizon and learns stably for all settings of $N$. However, the choice of $N$ has a marked effect on the speed of learning. Increasing $N$ speeds up learning up to a point after which no additional speedup is obtained (compare $N=0,1,5$). Importantly, these results suggest that the benefit of the model in IVG is not just one of representation learning in partially observed domains but that using the model for policy optimziation is beneficial.\\
\textbf{Averaging value gradients: } To quantify the effect of averaging value gradients across multiple rollout horizons (c.f. Eq \eqref{eqn:svgn_avg_full}), we ran experiments using a single imagined rollout of length $N$ ( \figref{fig:distractor_ablation}, right). For short horizons e.g. $N$=5, using a single rollout horizon leads to lower learning speed. On the other hand, learning completely fails for longer horizons $N$=20 (unlike with averaging) potentially due to cascading model errors on imagined rollouts, validating our averaging approach.

\vspace{-.2cm}
\section{Conclusions}
\vspace{-.2cm}
We presented an approach for model-based RL where an action-conditional latent space model is trained jointly with policy, value and reward functions that operate on the learned latent space. To achieve efficient policy optimization we introduced Imagined Value Gradients (IVG), an extension of SVG (using imagined rollouts and N-step horizon averaging). We demonstrated that IVG can learn complex, long-horizon manipulation tasks like lifting and stacking. We further demonstrated in several transfer experiments on related tasks that transferring a model learned via IVG can significantly improve data efficiency compared to off-policy baselines. Crucially, transferring with models trained on multiple tasks further accelerates learning, even succeeding on tasks where single-task transfer fails.
We feel that our approach is a promising first step towards designing RL methods that combine learning of closed loop policies with the generalization capabilities that learned approximate models can provide -- although extensions (such as handling egocentric cameras, increasing sample efficiency) are needed to make our approach a fully general purpose solution for real-world robotics tasks.

\acknowledgments{The authors would like to thank the entire Control Team and many others at DeepMind for numerous discussions on this work. Special thanks go to Tuomas Haarnoja and Raia Hadsell for reviewing an early version of this work and to Hannah Kirkwood for help in organizing the internship.}

\clearpage
%===============================================================================

\bibliography{references}  % .bib

%===============================================================================
\clearpage
%===============================================================================
\appendix

We provide additional details on some of the model components as well as full details regarding our experimental setup.

%%%%%%%%%%%%%%%%%%%%%%%%%%%%%%%%%%%%%%%%%%%
\section{Additional details on Value gradient derivation}
Given a model, we optimize a parametric policy $\pi_\theta(\bA | \bS)$ by maximizing the N-step surrogate value function which is a recursive composition of the policy, the transition, reward and value function:
\begin{equation}
    \VM_N(\bS^t) = \mathbb{E}_{\bA^k \sim \pi} \left[ \gamma^{N} \VE^\pi(\bS^{t+N}; \phi) + \sum_{k=t}^{t+N-1} \gamma^{k-t} \rE(\bS^k, \bA^k) \Big| \bS^{k+1} = f_\text{trans}(\bS^k, \bA^k) \right],
    \label{eqn_app:value_n}
\end{equation}
This N-step value can be computed by performing an ``imagined'' rollout in the latent state-space using our model (see Figure (1) in the main text).
It can be maximized by gradient ascent, exploiting the so called ``value gradient''  $\nabla_\theta{\VM}_N(\bS^t)$ \citep{heess2015learning}; which can often be computed recursively, taking advantage of the reparameterization trick \citep{kingma2013auto,rezende2014stochastic} for sampling from $\pi_\theta$ we can recursively define a sample estimate of this gradient. We start by defining the deterministic function $\pi_\theta(\bS^t, \epsilon)$ that transforms a sample $\epsilon$ from a canonical noise distribution $p(\epsilon)$ into a sample from $\pi_\theta(\bA | \bS)$.
In the following we will consider Gaussian policy distributions, i.e. $\pi_\theta(\bA | \bS) = \mathcal{N}(\mu_\theta(\bS), \sigma^2_\theta(\bS))$, for which the reparameterization is given as $\pi_\theta(\bS^t, \epsilon) = \mu_\theta(\bS) + \epsilon \sigma_\theta(\bS)$, with $p(\epsilon) = \mathcal{N}(\mathbf{0}, \mathbf{I})$, where $\mathbf{I}$ denotes the identity matrix. Using these definitions we can define the gradient $\nabla_\theta\VM_N(\bS^t)$ for any state as
% \begin{equation}
%  \nabla_\theta{\VM}_{N}(\bS^t) =  \sum_{k=t}^{t+N-1} \mathbb{E}_{p(\epsilon)} \Big[  \nabla_\theta \rE(\bS^{k}, \pi_\theta\big(\bS^{k}, \epsilon)\big) + \gamma \nabla_{\bS^{k+1}} {\VM}_{N-1+t-k}(\bS^{k+1}) \nabla_\theta f_\text{trans}\big(\bS^{k}, \pi_\theta(\bS^{k}, \epsilon)\big) \Big],
%  %\nabla_\theta{\VM}_{N}(\bS) = \mathbb{E}_{p(\epsilon)} \Big[  \nabla_\theta \rE(\bS, \pi_\theta\big(\bS, \epsilon)\big) + \gamma \nabla_{\bSp} {\VM}_{N-1}(\bSp) \nabla_\theta f_\text{trans}\big(\bS, \pi_\theta(\bS, \epsilon)\big) + \gamma \nabla_\theta{\VM}_{N-1}(\bSp) \Big],
% \label{eqn:svgn}
% \end{equation}

\begin{equation}
\nabla_\theta{\VM}_N(\bS^t) = \mathbb{E}_{p(\epsilon)} \big[\nabla_\theta \rE\big(\bS^{t}, \pi_\theta(\bS^{t}, \epsilon)\big) + \gamma \nabla_{\bS^\prime} \VM_{N-1}(\bS^\prime) \nabla_\theta f_\text{trans}\big(\bS^{t}, \pi_\theta(\bS^{t}, \epsilon)\big) + \gamma \nabla_\theta{\VM}_{N-1}(\bS^\prime) \big]
\label{eqn:svgn}
\end{equation}

%where we used the short-hand ``tick'' notation $\bSp = f_\text{trans}(\bS, \pi_\theta(\bS, \epsilon))$ 
where $\bS^\prime := \bS^{t+1} = f_\text{trans}(\bS^{t}, \pi_\theta(\bS^{t}, \epsilon))$
and we dropped dependencies of all functions on $\phi$ for brevity. The partial value gradient $\nabla_{\bS^\prime} {\VM}_{N-1}(\bS^\prime)$ wrt. a state $\bS^\prime$ is defined recursively as
% \begin{equation}
% \begin{aligned}
%  %\nabla_\theta{V}(N, \bS^t) \approx
%  \nabla_{\bS^k} {\tilde{V}}_{N}(\bS^k) = \mathbb{E}_{p(\epsilon)} \Big[ &\nabla_{\bS} \hat{r}(\bS^k, \bA\big) + \nabla_{\bA} \hat{r}(\bS^k, \bA)\big) \nabla_{\bS} \pi_\theta(\bS^k, \epsilon) +  \\
%  &\gamma \nabla_{\bS^{k+1}} \tilde{V}_{N-1}(\bS^{k+1}) \nabla_{\bS} f_\text{trans}\big(\bS^k, \bA\big) \big| \bA = \pi_\theta\big(\bS^k, \epsilon) \Big],
% %&+ \gamma \nabla_\theta \mathbb{E}_{\pi_\theta}\big[ V^\pi(\mathbf{s}^{t+1}; \bA^{t+1:N-1}) \big] \Big],  
% \end{aligned}
% \label{eqn_app:svgn_state}
% \end{equation}

\begin{align}
\nabla_{\bS^k} \VM_N(\bS^k) = \mathbb{E}_{p(\epsilon)} \Big[ & \nabla_{\bS^k} \rE\big( \bS^k, \bA^k \big) + \nabla_{\bA^k} \rE\big(\bS^k, \bA^k\big) \nabla_{\bS^k} \pi_\theta\big(\bS^k, \epsilon\big) +  \nonumber \\ 
& \gamma \nabla_{\bS^{k+1}} \VM_{N-1}\big(\bS^{k+1}\big) \nabla_{\bS^k} f_\text{trans}\big(\bS^k, \bA^k\big) + \nonumber \\ 
&  \gamma \nabla_{\bS^{k+1}} \VM_{N-1}\big(\bS^{k+1}\big) \nabla_{\bA^k} f_\text{trans}\big(\bS^k, \bA^k\big) \nabla_{\bS^k} \pi_\theta\big(\bS^k, \epsilon\big) \ \Big| \ \bA^k = \pi_\theta\big(\bS^k, \epsilon\big) \Big]
\label{eqn_app:svgn_state}
\end{align}

where the case $\nabla_\theta{\VM}_{1}(\bS^{t+N-1})$ is established by assuming that the policy does not change in any step after $N$; i.e. bootstrapping with $\nabla_{\bS^{k}}\VM_0(\bS^{k}) = \nabla_{\bS^{k}}\VE^\pi(\bS^{k}; \phi)$.
We note that, to calculate these gradients, only an initial state $\bS^t$ (encoded from a history of observations $\bO^{1:t} \sim \mathcal{B}$) is required in addition to the learned model. Our derivation here is thus analogous to the N-step stochastic value gradient definition from~\cite{heess2015learning} but replacing observed states with imagined latent states -- and assuming a deterministic transition model.

%%%%%%%%%%%%%%%%%%%%%%%%%%%%%%%%%%%%%%%%%%%

% \begin{figure}
%     \centering
%     \includegraphics[width=0.99\textwidth]{figures/training/policytraining.pdf}
%     \caption{Imagined rollout for the policy optimization. TODO: add more details}
% \end{figure}

\begin{figure}
    \centering
    \includegraphics[width=0.99\textwidth]{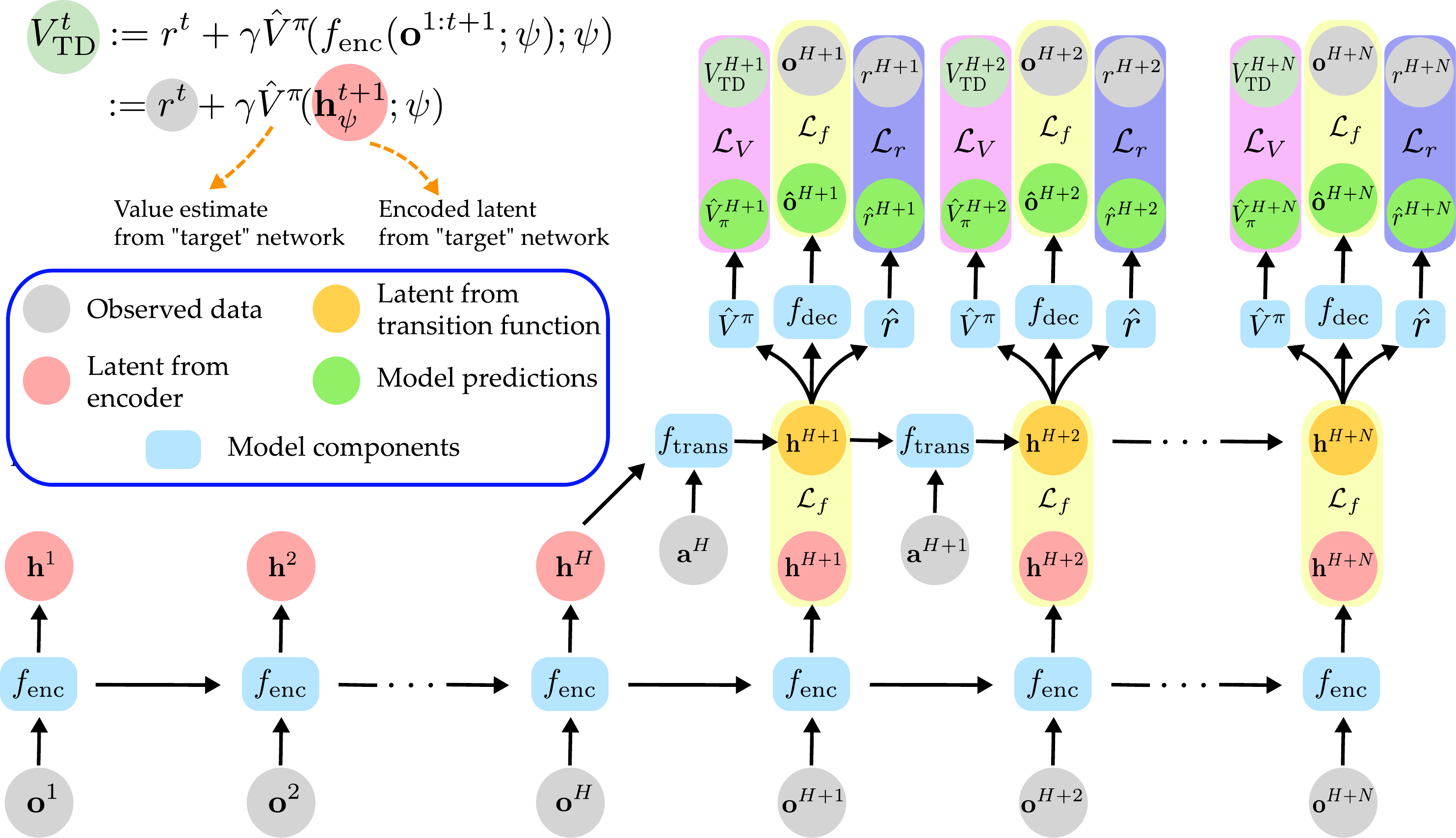}
    %\resizebox{14cm}{!}{\input{figures/modelrollout.tex}}
    \caption{Schematic showing the rollout from on a trajectory sampled from the replay buffer $(\bO^{1:H+N}, \bA^{1:H+N}, r^{1:H+N}) \sim \mathcal{B}$ through the model (blue rectangles), predicting latent states (red \& orange circles for encoder and transition predictions respectively) and their corresponding reconstructed observations, value and reward predictions (green circles). First, the encoder $f_{enc}$ encodes the observations $\bO^{1:H+N}$ to latents $\bS^{1:H+N}$. The open-loop rollout begins after a history of $H$ observations have been encoded to generate the latent $\bS^H$. From this latent, the rollout is computed through the transition model $f_\text{trans}$ using the true actions $\bA^{H:H+N-1}$, generating the latents $\bS^{H+1:H+N}$ (orange circles; note that these are different from the latents generated by the encoder). The transition model latents are passed through the decoder $f_\text{dec}$, value $\hat{V}^\pi$ and reward $\hat{r}$ estimators to generate (expected) reconstructed observations $\hat{\bO}^{H+1:H+N}$, (expected) values $\VE_{\pi}^{H+1:H+N}$ and (expected) rewards $\hat{r}^{H+1:H+N}$ which are used to compute losses for training the model (losses are highlighted with the rectangular color patches with the labels $\mathcal{L}_{(.)}$). Of special mention is the loss for training the value estimator; this uses V-trace and Temporal Difference (TD) style value targets based on a ``target'' network. The targets are generated by the ``target'' value estimator $\VE^{\pi}(. \ ; \psi)$ which takes in latents encoded by the ``target'' encoder $f_\text{enc}(. \ ; \psi)$ along with the observed rewards $r^t$. Best viewed in color.}
    \label{fig:modelactorrollout}
\end{figure}
\section{Additional details on Regularized Policy Optimization}
Given the definition of the value gradient in the previous section we can make a few interesting observations relevant to its use in practice. 

First, we can realize that, in principle, the single-step gradient estimate $\nabla_\theta{\VM}_1(\bS)$ (and hence a model trained by minimizing $\mathcal{L}^1$ is sufficient for performing policy optimization. However, in this case we would obtain a biased value gradient after one gradient step in the direction of $\nabla_\theta{\VM}_1(\bS)$ -- since at that point $\VE^\pi(\bS) \neq \VM^{\pi_\theta}(\bS)$ -- which only becomes unbiased again once the dynamic programming updates from \eqref{eqn_app:value_loss} have converged. 
To counteract this bias we could consider using the N-step gradient $\nabla_\theta{\VM}_N(\bS)$ for large $N$. Such an estimate is not affected by the above described bias for the first $N-1$ steps (since the equivalence of $\pi$ and $\pi_\theta$ is only assumed for steps after time $N$). As a result, it facilitates faster learning (as also demonstrated in our experiments). A downside of this approach is that it can be more heavily affected by modelling errors; i.e. a latent state-space model predicting rewards $N$-steps into the future is harder to learn than a 1-step model. As a compromise, trading-off bias with modelling errors, we found that using a simple average gradient estimate -- over $N$ horizons -- worked well in practice
$
    \nabla_\theta{\bar{V}}_N(\bS) = \frac{1}{N} \sum_{k}^N \nabla_\theta{\VM}_k(\bS).
    \label{eqn_app:svgn_avg}
$
This averaging linearly down weights the contributions from states further along the trajectory -- the first state appears N times in the sum, N-1 times for the second state and so on; as opposed to a discount based weighting that decays slowly this can drastically reduce the effect of model errors later in the sequence. We note that, in principle, we could also use weighting terms based on the variance of different horizon estimates, but opted for an average for simplicity here.

Second, even with the averaged model-gradient from above, gradient based optimization is prone to exploiting modelling errors (in both the transition dynamics and reward/value estimates), yielding overly optimistic policies. This is a well known problem in model based RL; see e.g. \citep{planet-dm} for a recent discussion. To counteract such effects it is hence desirable to further regularize the policy optimization step. Similar to many existing policy optimization methods \citep{schulman2017proximal,abdolmaleki2018maximum,haarnoja2018soft} we adopt a relative-entropy (KL) regularization scheme. We augment the estimated reward with a sample based likelihood ratio term (a sample based estimate of the KL)
\begin{equation}
    \rE_\text{KL}(\bS, \bA, \pi_\theta) = \rE(\bS, \bA) + \lambda \log\frac{\pi_\theta(\bA | \bS)}{p(\bA | \bS)},
\end{equation}
where $p(\bA | \bS)$ is a the prior action probability (we use $p(\bA | \bS) = \mathcal{N}(\mathbf{0}, \mathbf{I})$ throughout) and $\lambda$ is a multiplier trading-off reward and regularization. replacing $\rE$ in Equation \eqref{eqn:svgn} with $\rE_\text{KL}$ -- noting that $\rE_\text{KL}$ is differentiable wrt. the policy parameters $\theta$ -- results in the regularized value gradient 
% \begin{equation}
% \begin{aligned}
%  \nabla_\theta{V}^{\text{KL}}_{N}(\bS) = \mathbb{E}_{p(\epsilon)} \Big[ & \nabla_\theta r_\text{KL}(\bS, \pi_\theta\big(\bS, \epsilon), 
%  \pi_\theta \big) + \gamma \nabla_{\bSp} {V}^{\text{KL}}_{N-1}(\bSp) \nabla_\theta f_\text{trans}\big(\bS, \pi_\theta(\bS, \epsilon)\big) + \gamma \nabla_\theta{V}^{\text{KL}}_{N-1}(\bSp) \Big],
% \end{aligned}
% \label{eqn_app:svgn_reg}
% \end{equation}

\begin{equation}
\begin{aligned}
 \nabla_\theta{\VM}^{\text{KL}}_{N}(\bS) = \mathbb{E}_{p(\epsilon)} \Big[ & \nabla_\theta \rE_\text{KL}\big(\bS, \pi_\theta(\bS, \epsilon), \pi_\theta \big) + \\ 
 & \gamma \nabla_{\bSp} \VM^{\text{KL}}_{N-1}(\bSp) \nabla_\theta f_\text{trans}\big(\bS, \pi_\theta(\bS, \epsilon)\big) + \\
 & \gamma \nabla_\theta{\VM}^{\text{KL}}_{N-1}(\bSp) \Big],
\end{aligned}
\label{eqn_app:svgn_reg}
\end{equation}

where $\nabla_{\bSp}{\VM}^{\text{KL}}_{N-1}(\bSp)$ is, analogously, given by inserting $\rE_\text{KL}$ into Equation \eqref{eqn_app:svgn_state}. To ensure that the bootstrap value for ${\VM}^{\text{KL}}_{0}(\bS)$ is compatible with this regularized reward we additionally change the loss for the value function (the Bellman error); by, again, replacing $\rE$ with $\rE_\text{KL}$ in Equation \eqref{eqn_app:value_loss}. The total derivative estimate we use in practice is then given as
\begin{equation}
       \nabla_\theta \mathbb{E}_{\bar{p}} \Big[ \bar{V}^{\text{KL}}_N(\bS^t) \Big] \approx \mathbb{E}_{\bO^{1:t} \sim \mathcal{B}} \Big[ \frac{1}{N} \sum_{k=1}^N \nabla_\theta{\VM}^\text{KL}_k(f_\text{enc}(\bO^{1:t})) \Big],
    \label{eqn_app:svgn_avg_full}
\end{equation}
where we use batches of samples from the replay to optimize the policy on all visited states. That is we perform stochastic gradient ascent combining the gradient from Equation \eqref{eqn_app:svgn_avg_full} with any optimization method (we use Adam~\cite{kingma2014adam}). The full optimization procedure is also described in Algorithm.~\ref{alg:overall}.

\begin{algorithm}[t]
\caption{Imagined Value Gradients in Latent Spaces}
\label{alg:overall}
%\small
\begin{algorithmic} 
\STATE {Given: Empty experience dataset $\mathcal{B}$, burn-in $H$, rollout length $N$, episode length $T$}
\STATE \textbf{Each actor do:}
    \bindent
    %%%%%% Actor loop
    \WHILE{True}
    \STATE Fetch policy / model parameters ($\theta$, $\phi$) from learner; Initialize empty trajectory $\tau:=\emptyset$
    \FOR{t = 0 to T} 
    \STATE Apply control $\pi_{\theta}(f_{enc}(\bO^{t-H:t}; \phi), \epsilon), \ \epsilon \sim \mathcal{N}(\mathbf{0}, \mathbf{I})$ \COMMENT{Encode history, take action}
    \STATE Observe $r,\bO^\prime$; Insert ($\bO, \bA, r, \bO^\prime$) into $\tau$  
    \ENDFOR
    \STATE Save $\tau$ in the dataset $\mathcal{B}$
    \ENDWHILE
    \eindent

\STATE \textbf{Learner do:}
    \bindent
    \WHILE{True}
    %%%%%% Learner loop
    \STATE Sample sub-trajectory of length $H+N$ from buffer: $(\bO^{1:H+N}, \bA^{1:H+N}, r^{1:H+N}) \sim \mathcal{B}$
    \STATE Compute $\nabla_{\phi} \mathcal{L}^{\mathcal{N}}$, the gradient of Eq.~(3) w.r.t model, reward and value parameters $\phi$
    \STATE Compute the policy gradient $\nabla_\theta \mathbb{E}_{p_\pi} \Big[ \bar{V}^{\text{KL}}_N(\bS) \Big]$ (Eq.~(9)) w.r.t policy parameters $\theta$
    \STATE Grad ascent/descent: \ $\theta \leftarrow \text{Step}_\text{adam}(\theta, \nabla_\theta \mathbb{E}_{p_\pi} \Big[ \bar{V}^{\text{KL}}_N(\bS) \Big])$; \  $\phi \leftarrow \text{Step}_\text{adam}(\phi, -\nabla_\phi \mathcal{L}^N)$
    \ENDWHILE
    \eindent
\end{algorithmic} 
\end{algorithm}

\section{Details for the Value Learning step}
As described in the main paper value-loss $\mathcal{L}_V$ involves the calculation of a (squared) Bellman error, which is given by
\begin{equation}
\begin{aligned}
\mathcal{L}_{V}(\bS^t, \bA^t, r^{t}, \bO^{1:t+1}) = \frac{\pi(\bA^t | \bS^t)}{\mu(\bA^t | \bS^t)} \left(r^t + \gamma \VE^\pi(f_\text{enc}(\bO^{1:t+1}; \psi); \psi) - \VE^\pi(\bS^t; \phi)\right)^2,
\end{aligned}
\label{eqn_app:value_loss}
\end{equation}
where the next state value $\VE^\pi(f_\text{enc}(\bO^{1:t+1}; \psi); \psi)$ is calculated via a ``target network'', whose parameters $\psi$ are periodically copied from $\phi$, to stabilize training (see e.g. \cite{mnih2015human} for a discussion). In practice we use v-trace \citep{espeholt2018impala} to calculate a better target value. That is we set 
$\VE^\pi(f_\text{enc}(\bO^{1:t+1}; \psi); \psi) \hat{=} \VE_\text{trace}^\pi(f_\text{enc}(\bO^{1:t+1}; \psi); \psi)$,
where the v-trace target is given as:
\begin{equation}
    \VE_\text{trace}^\pi(\bS^t; \psi) = \VE^\pi(\bS^t; \psi) + \sum_{i=t}^{t+N-1} \gamma^{i-t} \Big(\Pi_{j=t}^{i-1} \rho_i \Big) \delta_i^V
\end{equation}
with $\delta_i^V = \rho_i (r_i + \gamma \VE^\pi(\bS_{i+1}; \psi) - \VE^\pi(\bS^t; \psi))$ being the temporal difference error multiplied by importance weight $\rho_i = \min\big(\rho_\text{clip}, \frac{\pi(a_i|s_i)}{\mu(a_i|s_i)}\big)$ with $\mu(a|s)$ denoting the behaviour policy and we set $\rho_\text{clip} = 1$. We refer to \citet{espeholt2018impala} for additional details regarding v-trace.
%%%%%%%%%%%%%%%%%%%%%%%%%%%%%%%%%%%%%%%%%%%
\section{Details for the Experimental Setup}
\label{sec:Appendix:Experiments}
%\akb{Got most of these details from the SAC paper. Check and rephrase. Maybe have someone else write it?}\\
We used the Mujoco Simulator\footnote{MuJoCo: see www.mujoco.org} for simulating the Sawyer robot setup. The robot is equipped with a two-finger Robotiq gripper. We ran the simulation with a numerical time step of 10 milliseconds, integrating 5 steps, to get a control interval of 50 milliseconds for the agent. In this way we can resolve all important properties of the robot arm and the object interactions in simulation.
All the objects used were based on wooden toy blocks and balls. For the majority of our experiments we used two cubic blocks with side lengths of 5 cm, colored red and blue. For the distractor experiments, we used a yellow cubic block of size 6 cm and a yellow ball of diameter 4 cm.
We used a table with sides of 60 cm x 30 cm in length as the workspace of the robot in all our experiments. Objects were spawned randomly on the table surface. The robot hand is initialized randomly above the table-top with a height offset of up to 20 cm above the table (minimum 10 cm) and the fingers in an open configuration. All experiments run on episodes with 200 steps length (which gives a total simulated real time of 10 seconds per episode). 

The sawyer robot is controlled via a 5D position controller based on an inverse kinematics model. That is, that agent outputs are 5D velocity commands -- 3 for the robot end effector position, one for the gripper and one for rotating the gripper to change its orientation. The action space for all our tasks, therefore, is 5 dimensional.

%%%%%%%%%%% Proprio / Image obs
\begin{table}[h!]
\centering

% Table 1 - proprio
\begin{tabular}{ l  c  c }
\hline
Entry & Dimensions & Unit \\
\hline
arm joint pos & 7 & rad \\
arm joint vel & 7 & rad / s \\
finger joint pos & 1 & rad \\
finger joint vel & 1 & rad / s \\
finger grasp & 1 & Binary \\
\hline \\
\end{tabular}
\quad
%
% Table 2 image obs
\begin{tabular}{ l  c  c }
\hline
Entry & Dimensions & Unit \\
\hline
camera 1 (front-left) & 3 x 64 x 64 & RGB \\
camera 2 (front-right) & 3 x 64 x 64 & RGB \\
\hline \\
\end{tabular}
\caption{\textit{Left}: Proprioceptive observations used in all simulation experiments. \textit{Right}: Image observations used in all simulation experiments (except the state based ones).}
\label{tbl_app:proprio_pixel}
\end{table}

%%%%%%%%%%% Feature obs
\begin{table}[h!]
\centering
\begin{tabular}{ l  c  c }
\hline
Entry & Dimensions & Unit \\
\hline
object i pose & 7 & m, au \\
object i velocity & 6 & m/s, dq/dt \\
object i relative pos & 3 & m \\
\hline \\
\end{tabular}
\caption{Object feature observations, used in the state based MPO experiment. Note that these are not used for any vision based experiment. The pose of the objects is represented as world coordinate position and quaternions. In the table m denotes meters, q refers to a quaternion which is in arbitrary units (au). $i$ denotes the id of the object; features from all objects are used as input.} %\akb{check}}
\label{tbl_app:features}
\end{table}

Table~\ref{tbl_app:proprio_pixel} (left) shows the list of proprioception observations we use for all our experiments. These observations are concatenated to produce a $17$ dimensional vector which is used as input to our model. In addition to proprioception, we use RGB images from two cameras located to the left and right of the table (in the front of the table, pointing towards the robot) as visual observations (see table~\ref{tbl_app:proprio_pixel}, right). These two images (3x64x64 each) are concatenated along the channel dimensions to generate a 6x64x64 input visual observation to our model. It is worth noting that the availability of two camera views helps disambiguate most occlusions. We will explore switching to a single central view in future experiments; this significantly increases occlusions and makes the tasks strongly partially observable. For the baseline state based MPO experiment, we used features of the objects (see table~\ref{tbl_app:features}) in addition to proprioceptive features as inputs to the policy.

~\figref{fig_app:recons_liftr} shows example images from a learned policy on the IVG task; both camera views used for training are shown.

\begin{figure}
    \centering
    \includegraphics[width=0.6\textwidth]{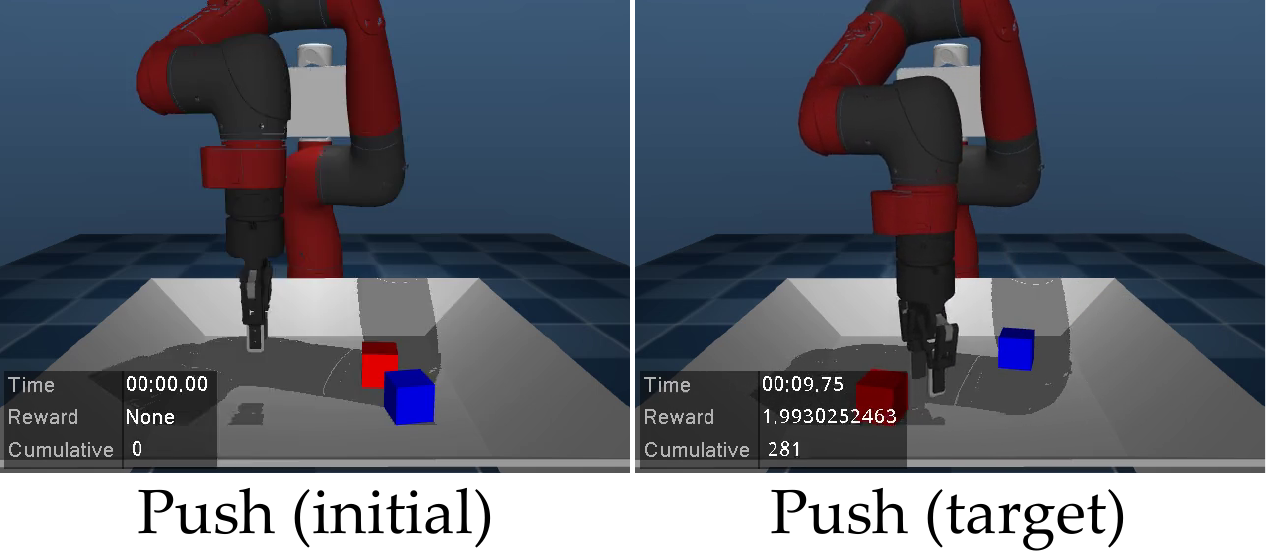}
    \caption{Example scenes from the match positions task. On the left is an initial image showing the blocks and the arm initialized to random starting positions. On the right, we show an image where the blocks are in their respective target positions (the blocks are always required to go to this target configuration in the task).}
    \label{fig_app:pushexample}
\end{figure}

\subsection{Tasks and Rewards}
We used shaped rewards for specifying all our tasks as we wanted to measure transfer efficiency rather than the capability to handle sparse reward settings. In principle, the multi-task version of IVG can be extended to the sparse reward setting easily, in a manner similar to SAC~\cite{riedmiller2018learning}. Below, we discuss the reward setup for the three major tasks considered in the paper, the \textbf{Lift}, \textbf{Stack} and \textbf{Match Positions} task. The addition of a \textit{visual distractor} does not change the reward setup for a given task.

\textbf{Lift: } In the lift task, the agent has to pick-up one of either the red (Lift-R) or blue (Lift-B) objects on the table and lift it above a certain height. We introduce additional shaping to the task through auxiliary rewards that encourage reaching the target object, grasping it and lifting it once grasped. These are specified in turn as:
\begin{itemize}  
\item\textit{REACH(O)}: $tol(d(TCP, O), 0.02, 0.15)$: \\
    Minimize the distance of the TCP to the target cube.
\item \textit{GRASP}: \\
    Activate grasp sensor of gripper ("inward grasp signal" of Robotiq gripper)
\item \textit{HEIGHT(O, x)}: $lin(O, x, 0.10)$ \\
    Increase z coordinate of an object more than $x = 0.03m$ relative to the table.
\end{itemize}
Where the $d(x,y)$ is the Euclidean distance between a pair of 3D points, and the tolerance and linear reward functions terms are defined as:
\begin{equation}
tol(v, \epsilon, r) =
\begin{cases}
  1 &\text{iff} \ |v| < \epsilon \\
  1 - tanh^2( \frac{atanh(\sqrt{0.95})}{r} |v|) &\text{else},
\end{cases}
\label{eq:shaped_tolerance}
\end{equation}

\begin{equation}
lin(v, \epsilon_{min}, \epsilon_{max}) =
\begin{cases}
  0 &\text{iff} \ v < \epsilon_{min} \\
  1 &\text{iff} \ v > \epsilon_{max} \\
  \frac{v - \epsilon_{min}}{\epsilon_{max} - \epsilon_{min}}  &\text{else}.
\end{cases}
\label{eq:shaped_tolerance2}
\end{equation}
The final reward is a weighted sum of all these sub-rewards: 
$$
LIFT(O) = REACH(O) + 0.5 * (GRASP + HEIGHT(O, 0.03)),
$$
which, overall cannot exceed a value of two. 

\textbf{Stack: } Similar to the lift task, there are two variants of the stack task: 1) Stack-R, where the agent has to stack the red block on the blue block and 2) Stack-B, where the agent does the opposite. We again introduce shaping by first encouraging the agent to lift the object -- the lift reward is a part of the reward for the stack task. Additionally, once the object has been lifted we encourage the agent to move towards the target, align it with the target block and release the grasped object. The total reward is: 
\begin{equation}
    STACK(O1, O2) = 
    \begin{cases}
      LIFT(O1) & \text{iff} \ HEIGHT(O1, 0.03) \leq 0.8 \\
      STACKED(O1, O2) & \text{else}.
    \end{cases}
\end{equation}
%$$
%STACK(O1, O2) = (HEIGHT(O1, 0.03) < 0.8) LIFT(O) + (HEIGHT(O1, 0.03) > 0.8) STACKED(O1, O2),
%$$
where:
$$
STACKED(O1, O2) = ABOVE(O1, O2) * NOTGRASP,
$$
where $NOTGRASP$ is detemined by the grasp sensor and:
$$ 
ABOVE(O1, O2) = tol(d(O1, O2), 0.02, 0.15) * HEIGHT(O1, 0.03)
$$

\textbf{Match Positions: } In this task, the agent has to move both the red and blue blocks to a fixed target position (see~\figref{fig_app:pushexample}). As this task involves moving both objects it is a nice setting for testing the generalization of our learned models. Additionally, the reward is not shaped to encourage motion towards an object; there is no change in the reward unless one of the objects is moved. We specify the reward as:
\begin{equation}
    MP(O1, O2) = tol(d(O1,t1),0.02,0.15) + tol(d(O2,t2),0.02,0.15)
\end{equation}
where $t1, t2$ denote the target 3D positions of the red and blue block respectively.

%Unlike the lift and stack tasks which are nicely shaped to encourage motion towards the objects, the reward for the match positions task is a mix of sparse and dense rewards. Th\\

\subsection{Partially observable environments}
For the experiments using more partially observable environments we considered two settings. In the first instance we added a delay to the proprioceptive features in the Stack-R task (see IVG(5) (delayed proprio) in Figure 3 in the main paper). This results in an environment where the RNN has to perform integration to estimate the robot arm position (and its velocities).

In the second experiment we created a variant of the Stack-R task in which the red block (that needs to be lifted and placed on top of the blue block) changes color (switching from blue to red at random every 2 frames). 

\subsection{Multi-task setup: Tasks and Rewards}
In the multi-task setup we introduce several auxiliary tasks that are solved in addition to a main extrinsic task. We consider the following tasks and rewards in all our multi-task experiments: \\
\begin{itemize}
\item $REACH(O_\text{Red})$
\item $REACH(O_\text{Blue})$
\item $LIFT(O_\text{Blue})$
\item $STACK(O_\text{Blue})$
\item $MOVE(O_\text{Blue})$
\end{itemize}
where the move reward is given by $MOVE(O) = tol(d(vel_O,v),3, 0)$ where $vel_O$ is the object's velocity.
To train in a multi-task setup we use a task-conditioned policy, value- and reward-function, we refer to the section below for details. The learned model (i.e. $f_\text{enc}, f_\text{trans}, f_\text{dec}$) on the other hand is not conditioned on the task -- it hence has to learn consistent dynamics across tasks. The actors generate data by selecting one task per episode at random.

We note that even though the multi-task largely consists of the same rewards as for individual experiments the data distribution is very different as the model is trained on episodes from all tasks. As can be seen, in the experiments in main paper, this results in significant improvements in the transfer learning experiments.

%%%%%%%%%%%%%%%%%%%%%%%%%%%%%%%%%%%%%%%%%%%
\section{Details on the Model \& Policy} \label{sec:modelarch}
\begin{figure}[h]
  \centering
    \includegraphics[width=0.9\textwidth]{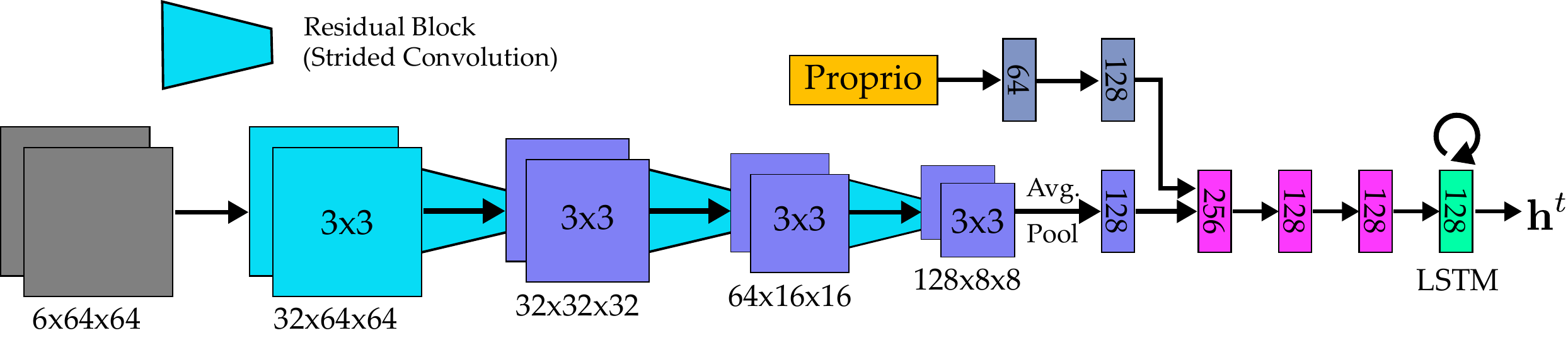}
  \caption{Network architecture of the encoder. The encoder takes in a pair of 64x64 RGB images concatenated along the channels axis and the proprioception observation and returns a 128-dimensional latent state vector ($\bS$) as output. It is implemented as a recurrent residual CNN with a final LSTM layer that integrates information across time.}
  \label{fig_app:encoder}
\end{figure}

We present some details on the architecture of the model components and policy network below:

The \textbf{encoder} $f_{enc}$ uses a recurrent, deterministic, convolutional neural network (CNN) to encode the observations $\bO^t$ to a low-dimensional latent state representation $\bS^t$ (see~\figref{fig_app:encoder}). Our observation is a pair of RGB images (3x64x64 each), concatenated along the first dimension, and a proprioception vector. The images are passed through a CNN with an initial convolutional layer followed by three residual blocks with strided convolutions~\cite{zagoruyko2016wide} and average pooling to generate a vector of outputs. In parallel, the proprioception input is passed through a 2-layer multilayer perceptron (MLP) to generate a feature vector. These are concatenated and passed through a 3-layer MLP and an LSTM which outputs the latent state $\bS^t$. As an initial pre-processing step, we normalized all our images to be between 0-1 and proprioception to -1 to 1. \\

\begin{figure}[h]
  \centering
    \includegraphics[width=0.65\textwidth]{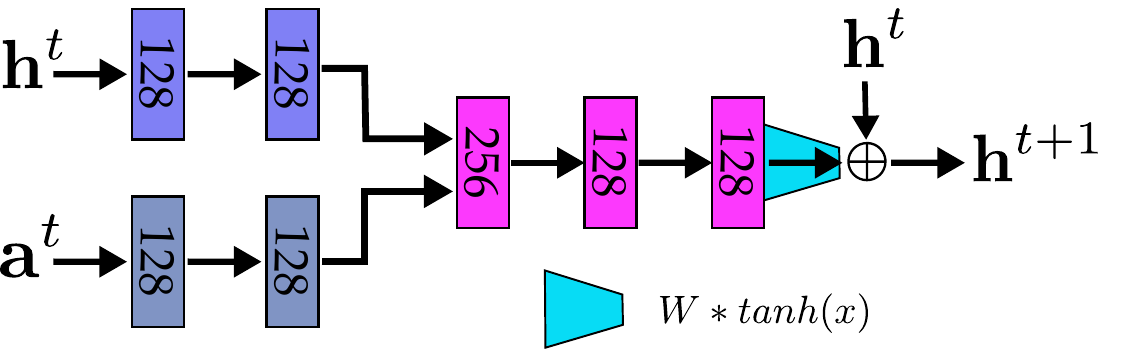}
  \caption{Network architecture of the transition model. The transition model takes a state ($\bS$) and action ($\bA$) as input and returns a prediction of the next state ($\bSp$). It is implemented as an MLP that predicts a delta change to the state ($\delta\mathbf{s}$) which is added to the input state to predict the output.}
  \label{fig_app:transition}
\end{figure}

The \textbf{transition} model $f_{trans}$ is deterministic, taking a latent state $\bS^t$ and action $\bA^t$ to predict the next latent state $\bS^{t+1}$ (see~\figref{fig_app:transition}). Both the inputs are first passed through 2-layer MLPs. The outputs of these MLPs are concatenated and passed through another 2-layer MLP which predicts the change in latent state $\delta{\bS}$. To ensure that the transition model outputs are well conditioned for long rollouts, we pass this delta change through a \textit{tanh} layer to normalize the result to -1 to 1. This is further scaled by a linear transform and added to the input state $\bS^t$ to generate the prediction $\bS^{t+1}$. In practice, we saw a significant improvement in performance when predicting the change in state as opposed to directly predicting the next state. %\akb{Do we want to talk about why deterministic?} \\

\begin{figure}[h]
  \centering
    \includegraphics[width=0.9\textwidth]{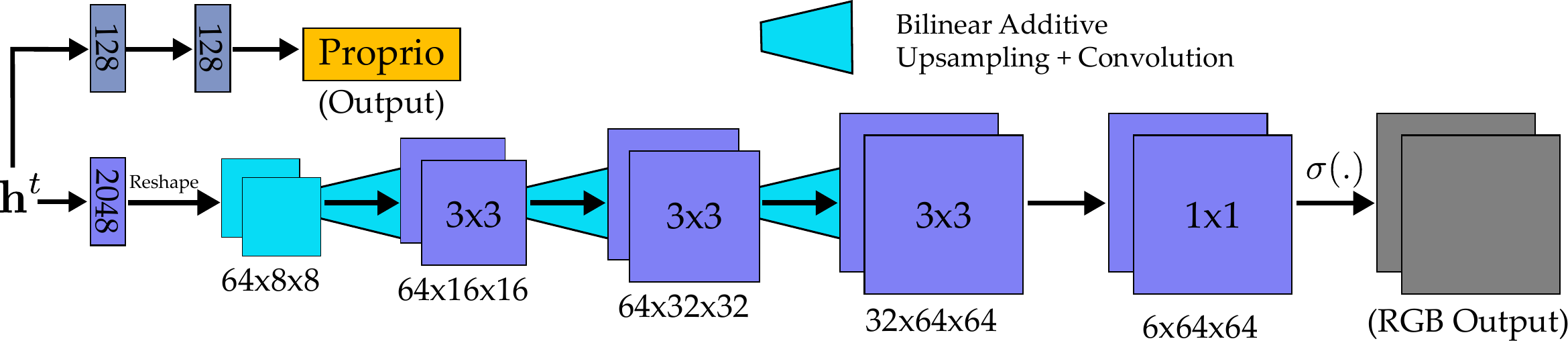}
  \caption{Network architecture of the decoder. The decoder takes a state ($\bS$) as input and returns a reconstruction of the corresponding RGB images and proprioception. We use an MLP to predict the proprioception output and a mix of bilinear upsampling and convolutional layers to generate the RGB reconstructions (which are normalized to 0-1 via a sigmoid).}
  \label{fig_app:decoder}
\end{figure}

The \textbf{decoder} $f_{dec}$ predicts the (expected) input observation $\bO^t$ from the latent state $\bS^t$ (see~\figref{fig_app:decoder}). We have two parts to the decoder: 1) To reconstruct the proprioception input, we pass the latent state through a 2-layer MLP. 2) For reconstructing the images, we first use a linear layer to transform the latent state to a $2048$ dimensional vector which is reshaped into a 64x8x8 feature tensor. This feature tensor is passed through three upsampling layers, each using a bilinear additive upsampling layer~\cite{wojna2017devil} followed by a convolution; the output is at the same resolution as the input images. Finally, the output features are passed through a 1x1 convolution layer to get the correct number of channels and a sigmoid layer to ensure that the outputs are normalized. We also experimented with using a de-convolutional architecture for the image upsampling but found that it reduced the reconstruction quality. \\

\begin{figure}[h]
    \begin{subfigure}{.5\textwidth}
        \centering
        \includegraphics[width=.95\linewidth]{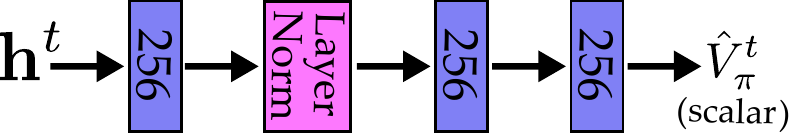}
    \end{subfigure}%
    \begin{subfigure}{.5\textwidth}
        \centering
        \includegraphics[width=.95\linewidth]{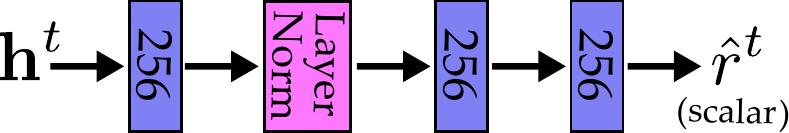}
    \end{subfigure}
    \caption{\textit{Left}: Network architecture of the value model $\VE^\pi$. The network takes the latent state $\bS^t$ as input and uses a three fully-connected layers to predict the expected value $\VE_{\pi}^t := \VE^\pi(\bS^t; \phi)$ (scalar). \textit{Right}: The reward model uses the same architecture as the value model but predicts the immediate reward $\hat{r}^t$ from the state $\bS^t$.}
    \label{fig_app:valuereward}
\end{figure}

The \textbf{value} $\VE^\pi$ and \textbf{reward} $\hat{r}$ modules predict the (expected) value $\VE_\pi^t := \VE^\pi(\bS^t; \phi)$ and reward $\hat{r}^t$ from a given state $\bS^t$ (see~\figref{fig_app:valuereward}). Both these modules are implemented as 3-layer MLPs, with a layer norm~\cite{ba2016layer} after the output of first layer. \\

Lastly, the \textbf{policy} $\pi_\theta(\bA | \bS)$ network predicts a distribution over actions $\bA^t$ from the corresponding latent state $\bS^t$. As mentioned earlier, we consider Gaussian policy distributions i.e. $\pi_\theta(\bA | \bS) = \mathcal{N}(\mu_\theta(\bS), \sigma^2_\theta(\bS))$, from which we can sample through the reparameterization trick as $\pi_\theta(\bS^t, \epsilon) = \mu_\theta(\bS^t) + \epsilon \sigma_\theta(\bS^t)$, with $p(\epsilon) = \mathcal{N}(\mathbf{0}, \mathbf{I})$, where $\mathbf{I}$ denotes the identity matrix. We implement the policy network as a 3-layer MLP similar to the \textbf{value} and \textbf{reward} modules. Unlike those, the policy outputs the mean $\mu_\theta$ and log-standard deviation $log(\sigma_\theta)$ from which we can sample an action using the reparameterization shown above.

\begin{figure}[h]
  \centering
    \includegraphics[width=0.8\textwidth]{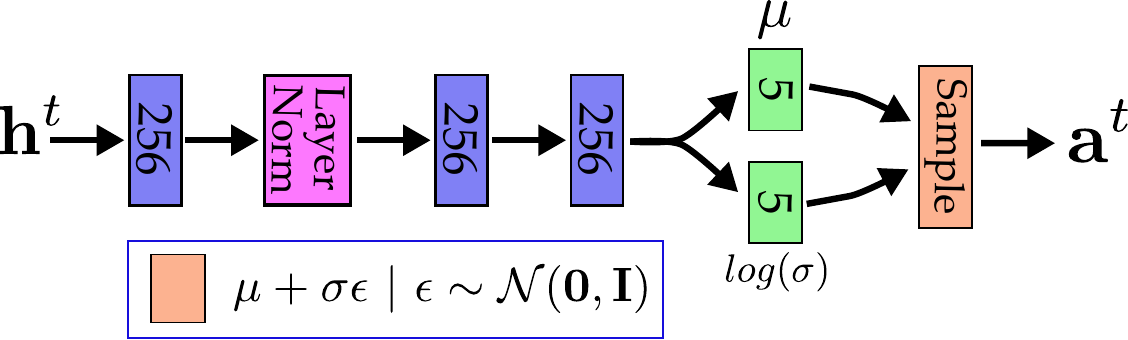}
  \caption{Network architecture of the policy $\pi_\theta(\bA|\bS)$. The policy takes as input the state $\bS^t$ and predicts the mean $\mu$ and log-variance $log(\sigma)$ of a Gaussian distribution over actions. We use the reparameterization trick to sample from this distribution by sampling $\epsilon \sim \mathcal{N}(\mathbf{0}, \mathbf{I})$. The output is a sampled action $\bA^t$.}
  \label{fig_app:policy}
\end{figure}

\subsection{Multi-task learning}
We introduce a few additional changes to the network architecture of a few model components and the policy in the multi-task learning setup. In this setting, each task has a unique \textbf{task ID} $\text{id}$ associated with it -- this is represented as a 1-hot vector of length $M$ ($M$ is the number of tasks). This task ID is fed as an additional input to both the \textbf{policy} ($\pi_\theta(\bA | \bS, \text{id})$) and \textbf{value} modules ($\VE^{\pi, \text{id}}$), thereby conditioning their predictions based on the task that is currently being considered. On the other hand, the \textbf{reward} predictor now predicts the rewards $\hat{r}^t$ for all these tasks; its output is now $M$-dimensional as opposed to a scalar from before. This further encourages the latent state to capture features relevant to all the learned tasks, leading to better generalization performance as witnessed in our experiments. Lastly, the architectures of the encoder $f_{enc}$, decoder $f_{dec}$ and transition model $f_{trans}$ are unchanged; these components are task agnostic and can integrate \& transfer knowledge across tasks.

%%%%%%%%%%%%%%%%%%%%%%%%%%%%%%%%%%%%%%%%%%%
\section{Details on training and transfer setup}
We implemented all our models in Python using the Tensorflow neural network package. Below, we present some details on the loss functions used for training and the hyper-parameter settings.

\subsection{Model Loss} 
We defined the per example model loss as $\mathcal{L}_e = \mathcal{L}_f(\bS^t, \bO^{1:t}) + \alpha \mathcal{L}_{r}(\bS^t, \bA^t, r^t) + \beta \mathcal{L}_{V}(\bS^t, \bA^t, r^{t}, \bO^{1:t+1})$. $\alpha$ and $\beta$ are 
coefficients that determine the relative contribution of the loss components. As explained in the main text, we use a squared error term for the reward loss $\mathcal{L}_{r}$ and a squared error to a V-trace target for the value loss $\mathcal{L}_{V}$. The per example transition model loss is given as
\begin{equation}
\begin{aligned}
\mathcal{L}_f(\bS^t, \bO^{1:t}) = \| f_{\text{dec}}(\bS^t; \phi) - \bO^t \|_2^2 + \zeta \|f_{\text{enc}}(\bO^{1:t}; \phi) - \bS^t \|_2^2,
\end{aligned}
\end{equation}
where the first term measures the error between the observations $\bO$ and reconstructions from the open-loop latent state predictions ($\bS^{t>H}$), the second term enforces consistency between the latent states predicted by the encoder $f_{\text{enc}}$ and the transition model $f_\text{trans}$ and $\zeta$ is a coefficient that determines the relative contribution of the two loss terms. The reconstruction loss is split into two parts (weighted equally), an image reconstruction loss and a proprioception reconstruction loss. We use a squared error term for the proprioception loss and a binary cross entropy loss term for image reconstruction; in practice we found this to result in better image reconstructions than a squared error term.

\subsection{Hyper-parameters} 
We used ADAM~\cite{kingma2014adam} with default settings and a fixed learning rate of 5e-5 
%\tobi{Checked correct}\akb{need to verify this} 
for all our experiments. We used the ELU~\cite{clevert2015fast} non-linearity as the activation function in all our networks. We initialized the final layers of our policy to predict values close to zero at the start of training; we found that this improved stability, especially in the early stages of learning. We used a latent state dimension of $N_S = 128$ for all our experiments ($|\bS| = 128$). We found this to be low-dimensional enough to be used for fast RL while still allowing room for expressivity.

We found that setting $\alpha = \beta = \zeta = 1.0$ gave the good results and kept this setting throughout all experiments. For the policy optimization, we set the weight of the KL regularizer to $\lambda = 0.01$ based on a hyper-parameter sweep. %\akb{check}

We used a batch size of $32$ (two learners each with a batch size of $16$) to train our model and policy in all our experiments; we initially experimented with larger batch sizes of $128$ but found that lowering the batch size made learning more stable. We fixed the history length $H=3$ and experimented with different rollout lengths $N=1,5,20$; as shown in our experiments $N=5$ performed best, we use that as the default. We ran experiments for a fixed number of episodes (per actor). %\akb{Check all this}

\subsection{Actor data generation}
We used $8$ asynchronous actors for data generation in all our experiments. At the start of each episode, the actor retrieves the most recent model and policy parameters. It then executes this policy a fixed time horizon of $T=10$ seconds (episode lasts 10 seconds). The resulting trajectory is split up into smaller sub-trajectories of length $H+N$, the length needed for learning, and added to a central replay buffer which collects experience from all actors. We used a buffer containing up to $100,000$ sequences (randomly deleting old sequences when full) for all our experiments. Both our learners sample from this replay buffer, compute the gradients for the model components and policy and perform synchronized updates to the parameters. %\akb{check}

For the multi-task experiments, at the start of each episode, the actor chooses a task to execute at random out of the $M$ available tasks. This task is executed for the full length of the episode ($T=10$ seconds). Random sampling of tasks can help generate diverse trajectories for training, facilitating learning of an expressive latent representation. %We also tried interleaving the execution of different tasks within each episode similar to the SAC setup \cite{riedmiller2018learning}; this performed significantly worse in practice.

\subsection{Baseline parameters}
%We used the following settings for the baseline algorithms:\\
%\akb{Tobi or Abbas: Can you fill this?}\\
For the baseline experiments that used pixel observations we constructed policy and value networks that are equivalent in architecture to applying $\pi_\theta$ after $f_\text{enc}$, similar to the networks in our IVG approach (to ensure a fair comparison).

For the state-based baselines we concatenated the true object positions, velocities and orientations (see table~\ref{tbl_app:features}) to the proprioceptive robot features (see table~\ref{tbl_app:proprio_pixel}, left). This is fed as input to 3-layer MLP policy networks (ELU activations, 200 hidden units each, layer normalization \citep{ba2016layer} after the first layer) and 3-layer MLP Q-value networks (ELU activations, 300 units each, layer normalization \citep{ba2016layer} after the first layer) which additionally take the actions (concatenated to other features) as input. To train SVG(0) we used the same relative entropy regularization technique as in for our method (using $\lambda = 10^{-3}$). For MPO we used the hyper-parameters from \citep{abdolmaleki2018maximum}, which performed well across all our tests. We tuned the learning rate for both MPO and SVG(0) for performance; a rate of 1e-4 worked best.

To ensure a fair comparison between algorithms in an asynchronous setting, we ensured all algorithms ran at the same frequency of learning steps per second (which we set to 10).

\subsection{Additional Baselines CEM and PG}
\paragraph{CEM}
To demonstrate the value of a parametric policy we ablate it and combine a model pre-trained with our approach with an implementation of the cross-entropy method (CEM). We bootstrap with a learned value estimate as in IVG (learning this value function in the transfer learning setting as in IVG) and perform latent rollouts.
In particular during training on a transfer task we replace the parametric policy with an optimization based approach using the cross-entropy method \cite{Rubinstein:2004:CEM:1014902}. We use CEM both for computiong actions in the value function learning step and when interacting with the system. In either case the length of the rollouts for CEM is set to 5 and we use 100 trajectories in each optimization step (repeating for a total of ten steps of optimization).

\paragraph{PG} For the likelihood ratio policy gradient baseline we use the exact same model and policy structure as for IVG. The only difference is in the calculation of the gradient of the state-action value (wrt. the policy parameters). In particular, we replace the value gradient from Equation \eqref{eqn:svgn_theta} with a likelihood ratio calculation \citep{Williams1992,Mnih2016a2c} (using 100 rollouts for the gradient estimation).
%This fails: CEM learns to reach for the object but then exploits the model, leading only to only ~60% lift success (transfer then does not result in any successful stacks). 

\subsection{Transfer experiment setup}
We use the following three-step procedure for transferring our models to new tasks:
\begin{enumerate}
    \item We first train the entire system from scratch (IVG) on a source task (or) a set of source tasks in the multi-task setting. From the trained modules, we choose the following model components: the encoder $f_{enc}$, transition model $f_{trans}$, and decoder $f_{dec}$. Only these components are transferred to the target task.
    \item We initialize the parameters of the encoder, transition model and decoder using the pre-trained networks. The parameters of the policy, value and reward functions are initialized to their default values; we train these from scratch.
    \item We train IVG in the usual fashion on the target task. An important point to note is that the encoder, transition and decoder networks are fine-tuned on the target task; they just have a significantly better initialization (that is generalizable to the target task). This is the primary contribution to our increase in learning speed in the transfer setting, particularly when using a model that has been trained on multiple source tasks.
\end{enumerate}

%\subsection{Setting up the exploration experiments}
%How we generate data from expert policy, add that to buffer, maybe make sure that expert data is a certain percentage in the buffer etc.

%%%%%%%%%%%%%%%%%%%%%%%%%%%%%%%%%%%%%%%%%%%
\section{Additional Experimental results}
We present a few additional experimental results in this section.
\subsection{Reconstructions}
\begin{figure}
    \centering
    \includegraphics[width=0.9\textwidth]{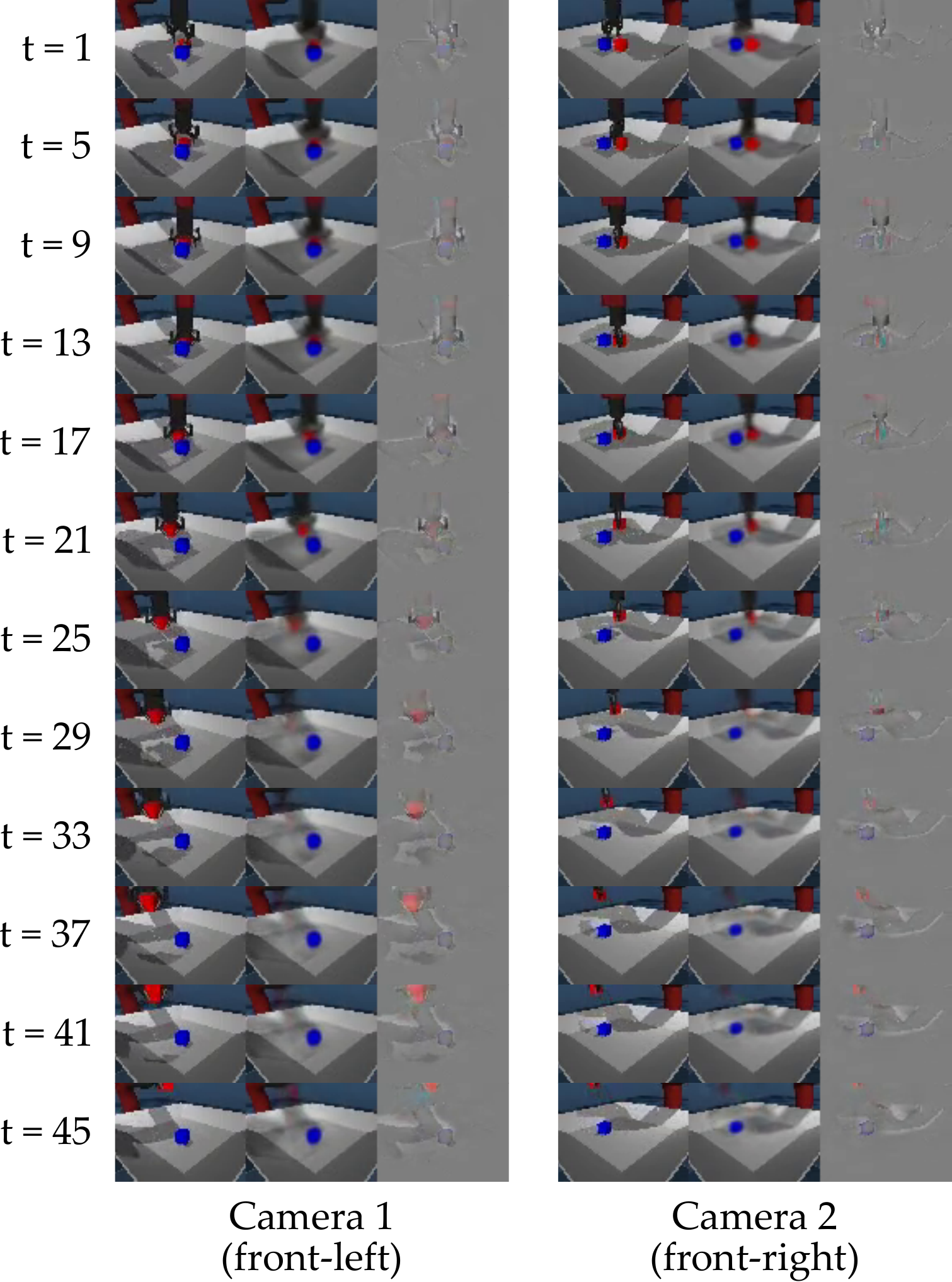}
    \caption{Example from the Lift-R task, showing open-loop reconstructions predicted by an IVG(5) model trained on the multi-task setting. The model gets a history of $H$ observations and encodes it into the latent state $\bS^H$ via the encoder $f_\text{enc}$. Next, it uses the sequence of actions $a^{H+1:H+N}$ to generate an open loop rollout through the transition model $f_\text{trans}$; this generates the transition states $\bS^{H+1:H+N}$. From these states, we run the decoder $f_\text{dec}$ to generate the image observations which are shown in this figure. \textit{Left}: Images from the left camera, reconstructions and error. \textit{Right}: Images from the right camera, reconstructions and error. For generating the predicted image sequence shown above, we set $H$=3 and $N$=45. The model was trained using IVG(5) i.e. $N$=5 in training. Even when running the model for significantly longer sequences than it was trained on, the predictions are consistent; the arm and objects are predicted well, albeit blurry, till around 30 frames. After 30 frames, the red object is poorly reconstructed (see the large errors near the object) but the arm positions and the blue block are still well predicted.}
    \label{fig_app:recons_liftr}
\end{figure}

\begin{figure}
    \centering
    \includegraphics[width=0.9\textwidth]{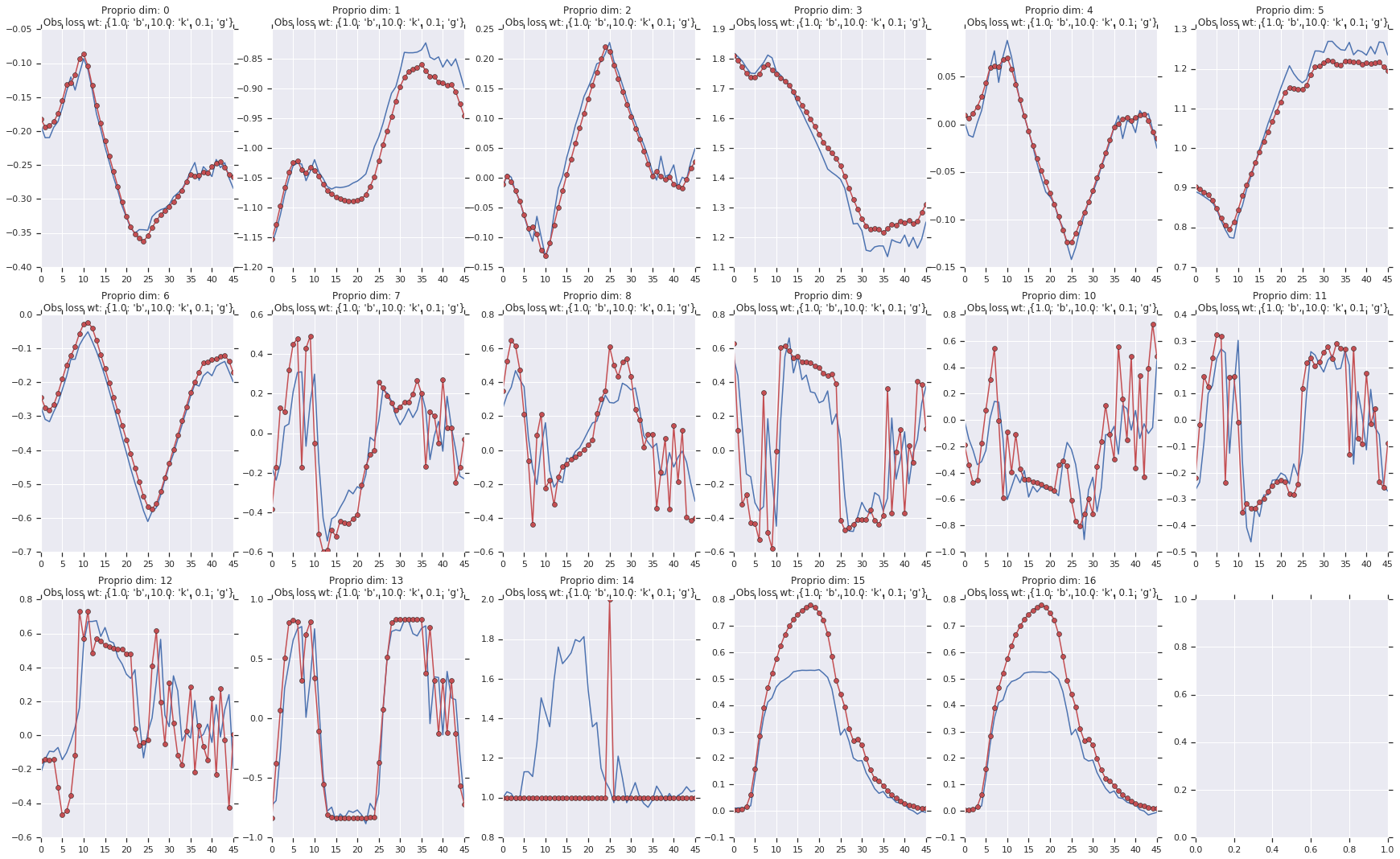}
    \caption{Example showing open-loop proprioception predictions from an IVG(5) model trained on the multi-task setting. Similar to the image reconstruction setting, we use a sequence of $H=3$ observations and an open loop sequence of $N=45$ actions ($\bA^{H+1:H+N}$) to generate the proprioception predictions (blue line). The targets are plotted in red. The predictions from the IVG model are largely consistent, even for dimensions that are highly noisy; the learned latent state is able to encode the proprioceptive features and the transition model can recover their dynamics well.}
    \label{fig_app:proprio_unknown}
\end{figure}

\figref{fig_app:recons_liftr} shows a 45-step open loop rollout from an IVG(5) model, trained on the \textit{Multitask} setting, from scratch. Even when tested on significantly longer sequences than it was trained on, the model predictions remain consistent, capturing salient details even after an open-loop prediction horizon of 30 frames. This also highlights an important point; even though the decoder predictions are not of high quality, the learned latent space can be used for robust control as it captures high-level task relevant details fairly well. Similarly,~\figref{fig_app:proprio_unknown} shows the result of a 45-step open loop prediction of proprioceptive features; the predictions are largely consistent with the observed results showing the strength of the learned model.

%TODO: Reconstructions, additional result plots w.r.t learner iterations maybe?\\ 
%\subsection{N-step gradient averaging}
%\tobi{Arun: add what you wanted here, not sure what goes here...}
%Talk about the linear downweighting which can help counteract cascading model errors -- maybe worth discussing this earlier when the value gradient is derived.

%TODO: Discuss empirical effect of N-step gradient averaging, large N slowdown, \\
%TODO: Stochastic envs, Distractors (maybe move one of the distractor plots to here if we get reward results)\\
%TODO: Adding demos for exploration tests\\

%\textit{Right}: Actor execution of the policy. The past $H$ observations are encoded to generate state $\bS^t$ which is fed to the policy to generate action $\bA^t$. This action is executed by the actor, generating observation $\bO^{t+1}$, and so on till the end of the episode.

\end{document}